\documentclass[letterpaper]{article} 
\usepackage{aaai25}  
\usepackage{times}  
\usepackage{helvet}  
\usepackage{courier}  
\usepackage[hyphens]{url}  
\usepackage{graphicx} 
\usepackage{natbib}  
\usepackage{caption} 
\usepackage{amsfonts}
\usepackage{epsfig}
\usepackage{graphicx}
\usepackage{amsmath}
\usepackage{amssymb}
\usepackage{comment}
\usepackage[ruled,vlined]{algorithm2e}
\usepackage{multirow}
\usepackage{booktabs}
\newcommand*\samethanks[1][\value{footnote}]{\footnotemark[#1]}

\usepackage{float}
\usepackage{newfloat}
\usepackage{listings}
\DeclareCaptionStyle{ruled}{labelfont=normalfont,labelsep=colon,strut=off} 
\floatstyle{ruled}
\newfloat{listing}{tb}{lst}{}
\floatname{listing}{Listing}
\title{Interacted Object Grounding in Spatio-Temporal Human-Object Interactions}
\author{
    Xiaoyang Liu\textsuperscript{\rm 1}\equalcontrib, 
    Boran Wen\textsuperscript{\rm 1}\equalcontrib, 
    Xinpeng Liu\textsuperscript{\rm 1,2}\equalcontrib, 
    Zizheng Zhou\textsuperscript{\rm 1}, 
    Hongwei Fan\textsuperscript{\rm 3}, \\ 
    Cewu Lu\textsuperscript{\rm 1},
    Lizhuang Ma\textsuperscript{\rm 1}, 
    Yulong Chen\textsuperscript{\rm 1}\thanks{Corresponding authors.},
    Yong-Lu Li\textsuperscript{\rm 1}\samethanks
}
\affiliations{
    \textsuperscript{\rm 1}Shanghai Jiao Tong University,
    \textsuperscript{\rm 2}Shanghai Innovation Institute,
    \textsuperscript{\rm 3}Peking University \\
    \{liuxiaoyang0309, xinpengliu0907\}@gmail.com, hwnorm@outlook.com, \\
    \{wenboran, zhou\_zz, lucewu, lzma, llong\_c, yonglu\_li\}@sjtu.edu.cn
}

\begin{document}

\maketitle
  
\begin{abstract}
Spatio-temporal Human-Object Interaction (ST-HOI) understanding aims at detecting HOIs from videos, which is crucial for activity understanding.
However, existing whole-body-object interaction video benchmarks overlook the truth that open-world objects are diverse, that is, they usually provide limited and predefined object classes. Therefore, we introduce a new open-world benchmark: \textbf{G}rounding \textbf{I}nteracted \textbf{O}bjects (\textbf{GIO}) including \textbf{1,098} interacted objects class and \textbf{290}K interacted object boxes annotation. 
Accordingly, an object grounding task is proposed expecting vision systems to discover interacted objects. 
Even though today's detectors and grounding methods have succeeded greatly, they perform unsatisfactorily in localizing diverse and rare objects in GIO. 
This profoundly reveals the limitations of current vision systems and poses a great challenge.
Thus, we explore leveraging spatio-temporal cues to address object grounding and propose a 4D question-answering framework (4D-QA) to discover interacted objects from diverse videos.
Our method demonstrates significant superiority in extensive experiments compared to current baselines. \textit{Data and code will be publicly available at \url{https://github.com/DirtyHarryLYL/HAKE-AVA}.}
\end{abstract}

\section{Introduction} 
As the prototypical unit of human activities, human-object interaction (HOI) plays an important role in activity understanding. 
Researchers begin with image-based HOI learning~\cite{hico-det,tin,idn,liu2022interactiveness,wu2022mining} and achieve great progress. 
Since daily HOIs require temporal cues to avoid ambiguity in detection, e.g., \textit{pick up-cup} and \textit{put down-cup}, video HOI task~\cite{epic-kitchen,daly,cad++,SomethingElse} is proposed to advance spatiotemporal HOI (ST-HOI) learning.

\begin{figure}
    \centering
    \includegraphics[width=\linewidth]{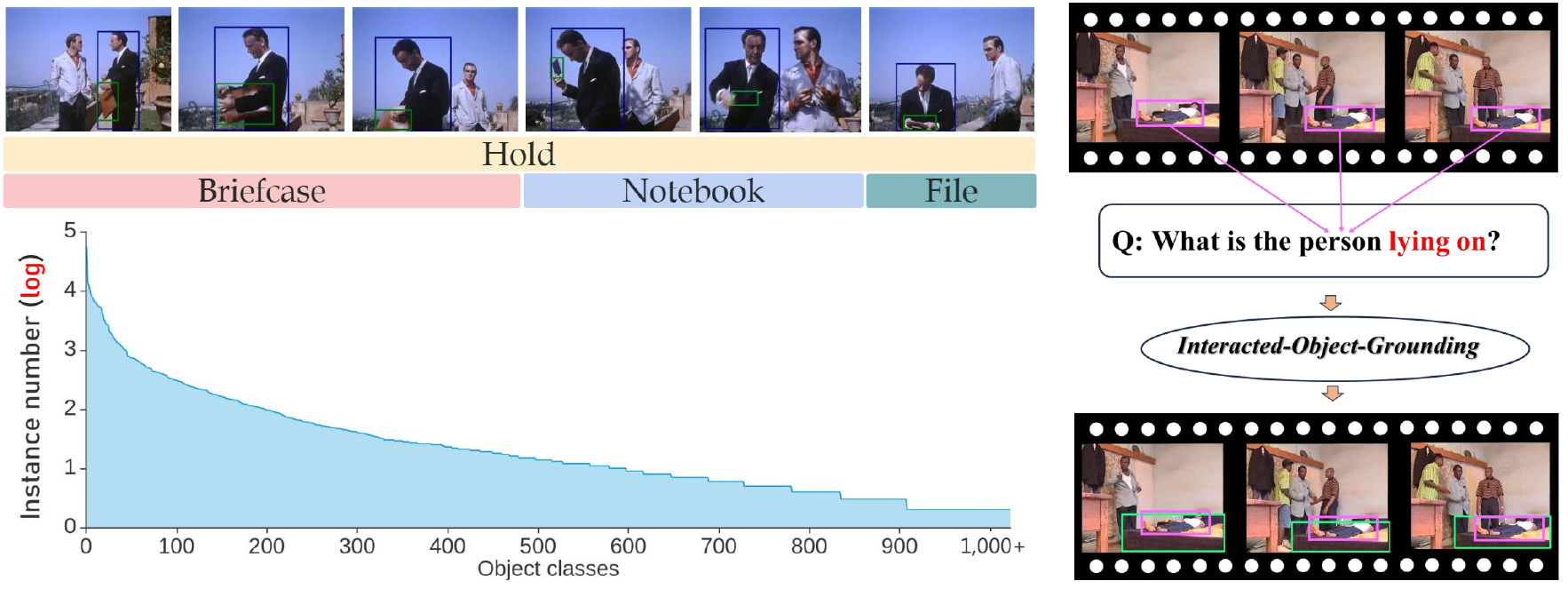}
    \caption{
    In daily HOIs, we interact with diverse objects with limited actions. 
    To this end, we build GIO on AVA~\cite{AVA}, annotating 1,000+ object classes to advance the study of HOI, with a long-tailed open-world object distribution. 
    We propose an open-world interacted object grounding task based on GIO as in the right figure. 
    Purple boxes indicate persons and green boxes indicate the grounded object.} 
    \label{fig:insight}
    \vspace{-20px}
\end{figure} 

However, many video HOI datasets are designed with limited predefined object classes.
Charades~\cite{charades}, DALY~\cite{daly}, Action Genome~\cite{ji2020action} all have less than 50 object classes (Tab.~\ref{tab:dataset_compare}). 
The limited object classes are less general for HOI tasks. 
Though some \textit{hand-object} interactions and \textit{egocentric} video-based HOI datasets include diverse objects like EPIC-Kitchens~\cite{epic-kitchen}, Something-Else~\cite{SomethingElse} and 100DOH~\cite{doh}, they focus on \textit{hand-object} interactions and \textit{egocentric} videos. 
As \textbf{whole body-object} interaction detection from \textbf{third-view} videos matters to numerous applications (e.g., health-care, security), here, we study third-person body-object interactions, such as \textit{ride/sit on} (chair, horse, etc), \textit{enter/exit} (train, bus, etc). 
Toward open-world HOI, we propose a large-scale third-view ST-HOI benchmark in this work, building upon AVA~\cite{AVA}: \textbf{G}rounding \textbf{I}nteracted \textbf{O}bject (\textbf{GIO}). 
It contains 1,098 interacted object classes within 51 interactions and 290K frame-level triplets $\langle human, verb, object \rangle$ as Fig.~\ref{fig:insight} shows.

Unlike previous works focussing on human/object tracking and action detection, we probed the complex ST-HOI through the \textbf{object view} given the largest scale of interacted object classes as in Fig.~\ref{fig:insight}.
We propose an open-world \textit{interacted object grounding} task with corresponding metrics to formulate this challenging problem. 
The initial formulation of ST-HOI(Sec.~\ref{sec:result}) suffers from \textbf{severe missing annotation}, which makes detection and evaluation less reliable.
Instead, our grounding task is insensitive to missing annotations, thus controlling the task's difficulty and reliability and enabling a meaningful analysis.
Given this task, cutting-edge image/video detectors~\cite{fasterrcnn,video-detector} fine-tuned on our train set all achieve less than 20 AP, even recent general visual grounding models based on large-scale VLMs~\cite{groundingdino} show limited performance.
Hence, GIO is still challenging and essential as the touchstone for open-world HOI.

Instead of directly regressing the object box, we devise a \textbf{4D question-answering (4D-QA)} paradigm.
First, the progress of the open-world segmentation model~\cite{sam} makes generating thorough and accurate fine-grained masks for arbitrary images possible. 
Then, a multi-option question-answering model is built to solve the problem: which masks correspond to the interacted object?
Multi-modal information is utilized to achieve this. Besides the raw video clip, we also reconstruct the 4D human-object layout for spatial clues and take it as a representation. Despite the pixel-level accuracy of the reconstruction is limited, it is sufficient for us to tackle the occlusion and spatial ambiguities for object localization. In comparison to directly regressing the object box, the 4D human-object layout before the QA paradigm provides general object-orient HOI information, this is why our method can achieve significant improvement. 
We believe GIO would inspire a new line of studies and pose new challenges and opportunities for the development of deeper activity understanding.

Our contributions are three-fold:
(1) We probe ST-HOI learning via an interacted object view and build a large-scale third-view ST-HOI benchmark GIO, including 290K open-world interacted object boxes from 1,098 object classes. 
(2) A novel interacted object grounding task is proposed to drive the studies on finer-grained activity parsing and understanding. 
(3) Accordingly, a 4D question-answering framework is proposed and achieves decent grounding performance on GIO with multi-modal information. 

\section{Related Works}
\label{sec:related-work}

{\bf Object Tracking.}
Object tracking is an active field and has two main branches, \textit{i.e.}, Single-Object Tracking~\cite{siamban,fan2019lasot} and Multi-Object Tracking~\cite{ristani2016performance,braso2020learning}.
Recently, tracking-by-detection~\cite{kim2015multiple,sadeghian2017tracking} has received lots of attention and has achieved state-of-the-art performance.

\noindent{\bf Human-Object Interaction (HOI).}
In terms of image-based HOI learning, both image-level~\cite{hico,pastanet,kato2018compositional} and instance-level~\cite{hico-det,tin,havev1,tinpami,djrn,liu2022highlighting} methods achieve successes with the help of large-scale datasets~\cite{hico-det,pastanet}.  
As for HOI learning from third-view videos, recently many large-scale datasets~\cite{AVA,charades,ji2020action,doh,vlog,activitynet} are released to promote this field, thus providing a data basis for us. 
They provide clip-level~\cite{activitynet,vlog,charades} or instance-level~\cite{AVA,ji2020action,daly} action labels, but few of them afford diverse object classes. 
Though some datasets~\cite{SomethingElse,epic-kitchen} provide instance labels of diverse object classes, they usually concentrate on egocentric hand-object interaction understanding~\cite{xu2022learning}.
Relatively, we focus on whole-body-object interaction learning based on third-view videos and propose GIO featuring the discovery of diverse objects.
Recently, there are also methods studying video-based visual relationship~\cite{vidvrd,liu2020beyond} and HOI~\cite{gpnn,strg,baradel2018object,videotransformer}.

\noindent{\bf Object Detection and Localization.} 
Object detection~\cite{fasterrcnn,yolo} achieves huge success with deep learning and large-scale datasets~\cite{coco} but may struggle without enough training data.
Some works~\cite{fsod} study few/zero-shot detection.
Moreover, as videos can provide temporal cues of moving objects, video object detection~\cite{video-detector} also received attention.
Unlike typical detection, some studies try to utilize context cues, such as human actor~\cite{kim2020tell,gkioxari2018detecting}, action recognition~\cite{yuan2017temporal,yang2019activity}, object relation~\cite{hu2018relation}, to advance object localization. 
\citet{gkioxari2018detecting} treated object localization as density estimation and used a Gaussian function to predict object location.
\citet{kim2020tell} borrowed human pose cues and language prior, constructing a weakly-supervised detector.
Moreover, object grounding with language descriptions also attracts attention in the vision-language crossing field, with promising potential in open-vocabulary object detection.
\citet{li2022grounded} formulates object detection as an object grounding problem for open-vocabulary object detection. 
\citet{yao2022detclip} boosted data from image captioning datasets for generalization ability.
\citet{groundingdino} extended the powerful DINO~\cite{zhang2022dino} model for the object grounding pipeline, achieving impressive performance.
\citet{sadhu2020video} grounded objects in video clips given language descriptions.

\section{Constructing GIO}
\label{sec:datset}

\subsection{Data Collection}

\begin{table*}[t]
\centering
\caption{Dataset comparison. 
Instances/triplets are in \textit{frame-level}.
\textit{18K*}: object class labels of \citet{SomethingElse} are uncurated.
In \textit{Action Genome*} and \textit{VidHOI*}, spatial relationships are not regarded as HOI.}
\vspace{-10px}
\resizebox{.9\linewidth}{!}{
\setlength{\tabcolsep}{3.5pt}
\begin{tabular}{l ccc cc cc cc}
\hline 
 \multirow{2}{*}{Dataset}  & \multirow{2}{*}{Video Hours} & \multirow{2}{*}{Annotated Frames} & \multicolumn{2}{c}{Objects} & \multicolumn{2}{c}{HOI} & \multirow{2}{*}{HOI/frame} &\multirow{2}{*}{View} &\multirow{2}{*}{Subjective}\\ 
 \cline{4-5}\cline{6-7}
 & & & class & instance & class & triplet & \\ 
\hline
Something-Something~\cite{goyal2017something} & 121 & 108K  & - & - & 174 & - & - &first & hand\\
100DOH~\cite{doh} & 3144 &  100K & - & 110.1K & 5 & 189.6K & 1.90 & first, third & hand\\
Something-Else~\cite{SomethingElse} & - & 8M & 18K*
& 10M & 174 & 6M & 0.75 & first & hand\\
EPIC-Kitchens~\cite{epic-kitchen}  & 55 & 266K & 331 & 454K & 125 & 243K & 0.91 & first & hand\\ 
CAD120++~\cite{cad++} & 0.57 & 61K & 13 & 64K & 10 & 32K & 0.52 & third & head, hand\\
VLOG~\cite{vlog}  & 344 & 114K & 30 & - & 9 & - & -  & first, third & hand\\ 
\hline
AVA~\cite{AVA} & 107 & 351K & - & - & 51 & - & - & third & whole body\\  
Charades~\cite{charades} & 82 & 66K & 46 & 41K  & 30 & - & - & third & whole body\\
DALY~\cite{daly} & 31 & 11.9K & 43 & 11K & 10 & 11K & 0.92 & third & whole body\\
Action Genome~\cite{ji2020action}* & 82 & 234K(227K*) & 35 & 476K & 15* & 454K* & 2.01 & third & whole body\\
VidHOI~\cite{chiou2021st}* & 70 & 217K(146K*) & 78 & - & 39* & 278K* & 1.90 & third & whole body \\
\hline
GIO & 74 & 126K & \textbf{1,098} & 290K & 51 & 290K & \textbf{2.30} & third & whole body\\
\hline
\end{tabular}}
\label{tab:dataset_compare}
\vspace{-15px}
\end{table*}
To support practical ST-HOI learning, we collect third-view videos from large-scale dataset AVA~\cite{AVA}. It contains 430 videos with spatio-temporal labels of 80 atomic actions (body motions and HOIs). As AVA includes complex HOIs in diverse scenes, it can bring great visual diversity to our benchmark.
We extract the HOI-related frames and the corresponding human boxes and action labels, thus the clips in GIO have uneven temporal durations. Notably, we only consider the non-human objectives in HOIs. 
Overall, based on the available train and validation (val) sets of AVA 2.2~\cite{AVA} (299 videos), we chose 74 hours of video including 51 actions (detailed in the supplementary).

\subsection{Dataset Annotation}
AVA provides labels with a stride of 1s, so we add boxes and class labels for all interacted objects with the same stride. Following AVA, we define the annotated frame as key frames which are at 1-second intervals.

\textbf{First}, as humans can perform multi-interaction simultaneously, we set the \textbf{annotating unit} as a clip including one \textit{single} interaction to normalize the annotation.
For example, a 30s clip including an actor \textit{holds-sth} (1-30s) and \textit{inspects-sth} (10-15s), will be divided into two sub-clips, i.e., a 30s sub-clip for \textit{holds-sth} and a 5s sub-clip for \textit{inspects-sth}. 
In brief, each sub-clip contains \textbf{one} verb and \textbf{one/several} class-agnostic interacted objects. 
Then, sub-clips are annotated separately, and each one is annotated by at least 3 annotators and checked by an expert to ensure quality.

\textbf{Second}, as AVA contains various scenarios and diverse objects, to better locate objects and avoid ambiguity, each annotator is given a whole sub-clip to draw boxes and classify them. 
In default, we use COCO~\cite{coco} 80 objects as a class pool. If annotators think an object is not in the pool, they are asked to input a suitable class according to their judgments. If an object cannot be recognized, they can choose the ``unknown'' option. Then, we find that a surprising 42.66\% of object instances are beyond our pool. After exhaustive annotation, we fix the input typos, exclude outliers via clustering, and combine similar items. Finally, 1,098 classes are extracted after cleaning. We then conduct re-recognition for the frames including ``unknown'' objects. 

\textbf{Finally}, to generate the ST-HOI labels, we further consider the objects in each sub-clip (one interaction of one person).
If there is only one object in a sub-clip, we use its locations as the labels.
If there are multiple objects, we record all of their boxes and manually link their boxes as multiple-object tracklets.
Then, each sub-clip is seen as a \textit{ST-HOI traklet}, whose label records \textit{a human actor tracklet, an interaction, a/several class-agnostic object tracklets}.

\subsection{Dataset Statistics and Attributes}
\label{sec:datastatics}
GIO includes 290K HOI triplets and 290K object boxes of 1,098 classes, including a wide range of rare objects. 
Only \textbf{20.85}\% of our object classes are covered by the recent large-scale object dataset FSOD~\cite{fsod}. 
It is noteworthy that Action Genome and VidHOI include predicates such as \texttt{next to}, which are not HOIs. Consequently, we refined the annotations and recalculated the statistics in Tab.~\ref{tab:dataset_compare}. In contrast, GIO, aiming for diversity and finer granularity, offers the highest number of object classes and the richest HOI instances per frame (\textbf{2.30}).

\subsection{Interacted Object Grounding}
GIO supports ST-HOI detection and many fine-grained tasks, like object classification. However, the original ST-HOI task, involving detection, tracking, and action recognition, is highly complex and challenging, with most approaches facing significant difficulties due to the task's inherent complexity and the quality of annotations. So instead of requiring vision systems to detect complete ST-HOI triplets, we focus on GIO's capability for interacted object grounding, i.e., given the human actor tracklet (and the interaction semantics), while object labels are not included in the interaction semantics, probing the ST-HOI understanding from the object view.
To make our task realistic, 328 object classes only have less than 5 samples (boxes) in our train set, and 98 classes are \textit{unseen} in the inference.

\section{Method}
\begin{figure*}[!t]
\vspace{-10px}
	\begin{center}
        \centerline{\includegraphics[width=.8\linewidth]{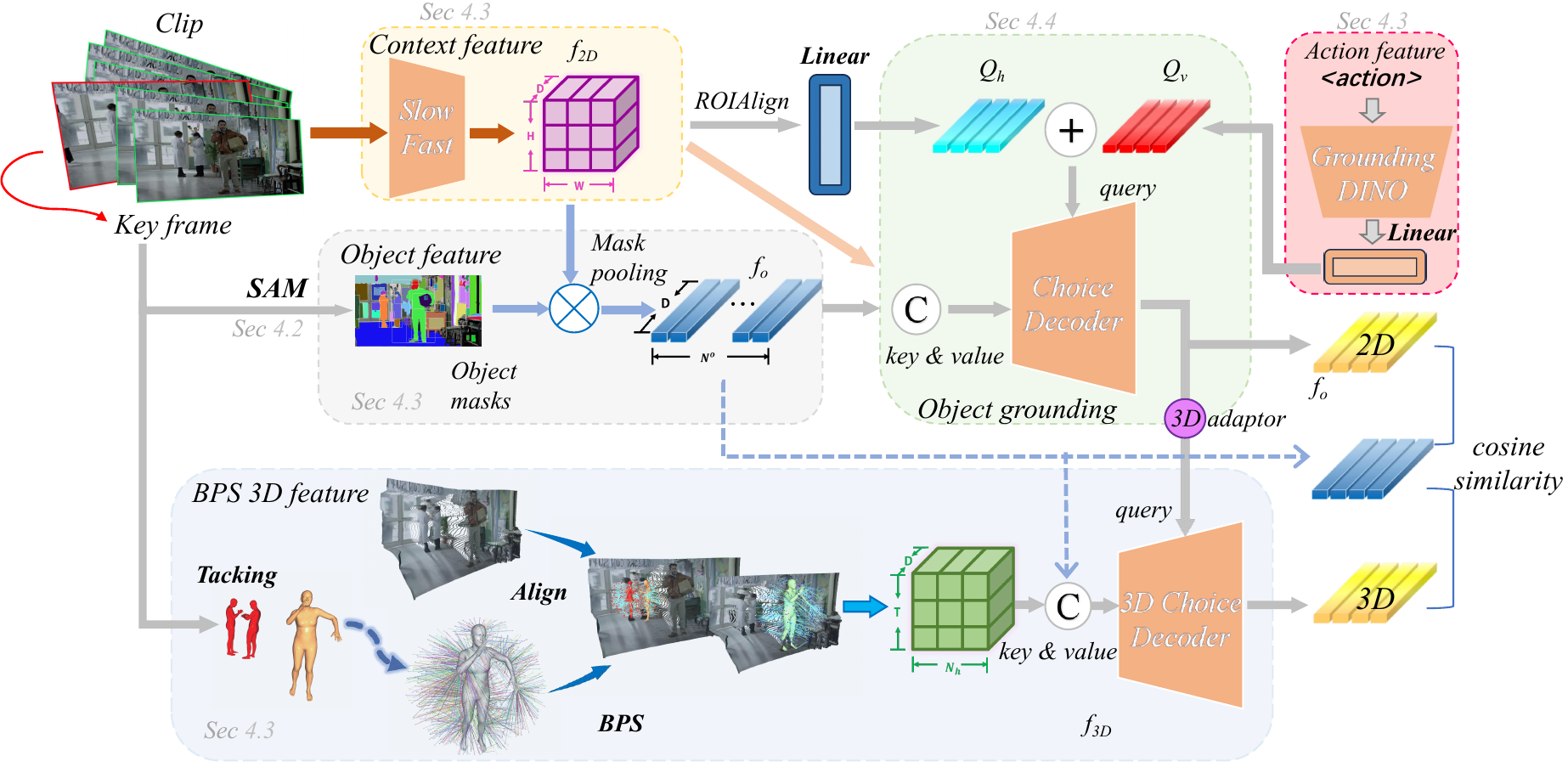}}
	\end{center}
        \vspace{-25px}
	\caption{The overview of our 4D-QA. It utilizes a 4D question-answering paradigm to effectively locate the interacted objects.} 
    \label{fig:HPN} 
    \vspace{-15px}
\end{figure*}
\label{sec:method}
In this section, we describe the pipeline of our method (Fig.~\ref{fig:HPN}). 
We focus on interacted object grounding, i.e., given the human actor tracklet (and the interaction semantics), systems are required to ground the interacted object. 
The difference between our task and the common object grounding tasks is our focus on the specific interaction between the grounded object and the person (\textit{interactiveness}), which makes it more difficult.
For clarity, the description unit hereinafter is \textit{one human tracklet} including one tracked person.  

\subsection{Overview}
Given a clip $C$, the target human tracklet $T_h=\{I^k_h\}^n_{k=1}$ ($n$ for tracklet length), we aim at learning a model $\mathcal{M}$ as 
\begin{equation}
    \hat{T_o} = \mathcal{M}(C, T_h, \{s, \emptyset \}), 
\end{equation}
where $\hat{T_o}$ is the predicted interacted object tracklet and the interaction semantics $s$ is an \textit{optional} input to inform the system with high-level semantics. 
To achieve this, instead of directly regressing the object box, we adopt a novel 4D question-answering paradigm to leverage HOI prior.
Given the strong generalization ability of SAM~\cite{kirillov2023segany}, we adopt it as an objectness detector to generate candidate object proposals (Sec.~\ref{sec:sam}).
The clip $C$ is first fed to SAM, resulting in $K$ candidate object mask tracklets $M_o=\{M_o^i\}_{i=1}^K$.
The task is then reformulated as choosing the interacted mask tracklets from the candidate tracklets, as 
\begin{equation}
    \hat{T_o} = \mathcal{M}(C, T_h, \{s, \emptyset \}, M_o).
\end{equation}

To tackle the challenging GIO, a 4D question-answering network is devised as shown in Fig.~\ref{fig:HPN}.
Multimodal features, including 4D clues, are extracted in the inspiration of DJ-RN~\cite{djrn}. 
We begin by extracting spatiotemporal features from the video using the SlowFast~\cite{feichtenhofer2019slowfast} network as a basis. 
Then, the 4D Human-Object layout is reconstructed for feature extraction (Sec.~\ref{sec:MM}).
Finally, we ground the interacted object with two decoders to summarize the important clues in complex spatiotemporal patterns (Sec.~\ref{sec:decoder}).
Despite the suboptimal precision of 4D Human-Object reconstruction, it is effective in alleviating the \textit{view ambiguity} in clips, also enhancing the object localization with 3D spatial information. The question-answering paradigm eases the learning process.

\subsection{SAM-based Candidate Generation}
\label{sec:sam}
We chose SAM as the candidate proposal generator for several reasons. 
First, SAM, based on pixel-level segmentation, provides a finer granularity and more accurate segmentation. 
Second, AVA consists of many video scenes that are dark, complex, and contain numerous objects. 
Traditional detection methods struggle to accurately predict small and blurry objects in such challenging scenarios. 
In contrast, SAM's pixel-based segmentation is more robust and accurate than directly predicting object bounding boxes. 
In addition, SAM is also adept at dealing with large objects. 
However, SAM could segment objects into multiple parts. 
Thus, our policy is to predict vast majority of the masks belong to the object resulting in a highly accurate bounding box.

{\bf Mask proposal generation.} 
Given a clip $C$, we denote the keyframe as $C_k$.
SAM is first fed with a grid of point prompts on $C_k$.
Then, low-quality and duplicate masks are filtered out. 
As a result, each image would produce at most 255 masks as $M_o$, which will be sent to the model as proposals to generate the final object box. 

{\bf GT proposals.} 
To judge which mask is GT, we input the GT object box to SAM as the prompt to get an accurate mask ($M_{acc}$). 
\begin{figure}[!t]
\centering
    \begin{minipage}{.45\linewidth}
        \centering
        \includegraphics[width=\linewidth]{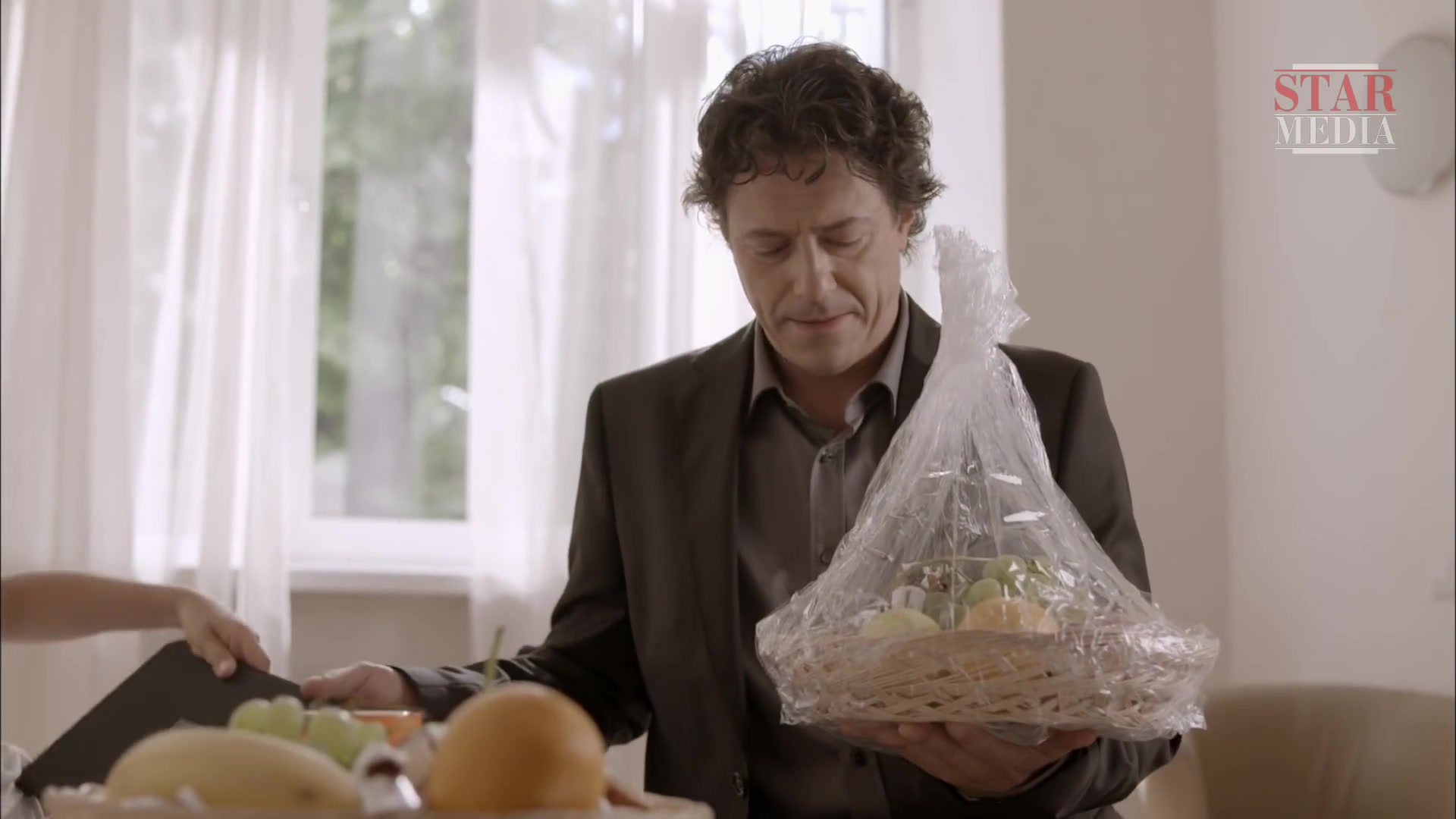}
        (a) Original image.
    \end{minipage}
    \begin{minipage}{.45\linewidth}
        \centering
        \includegraphics[width=\linewidth]{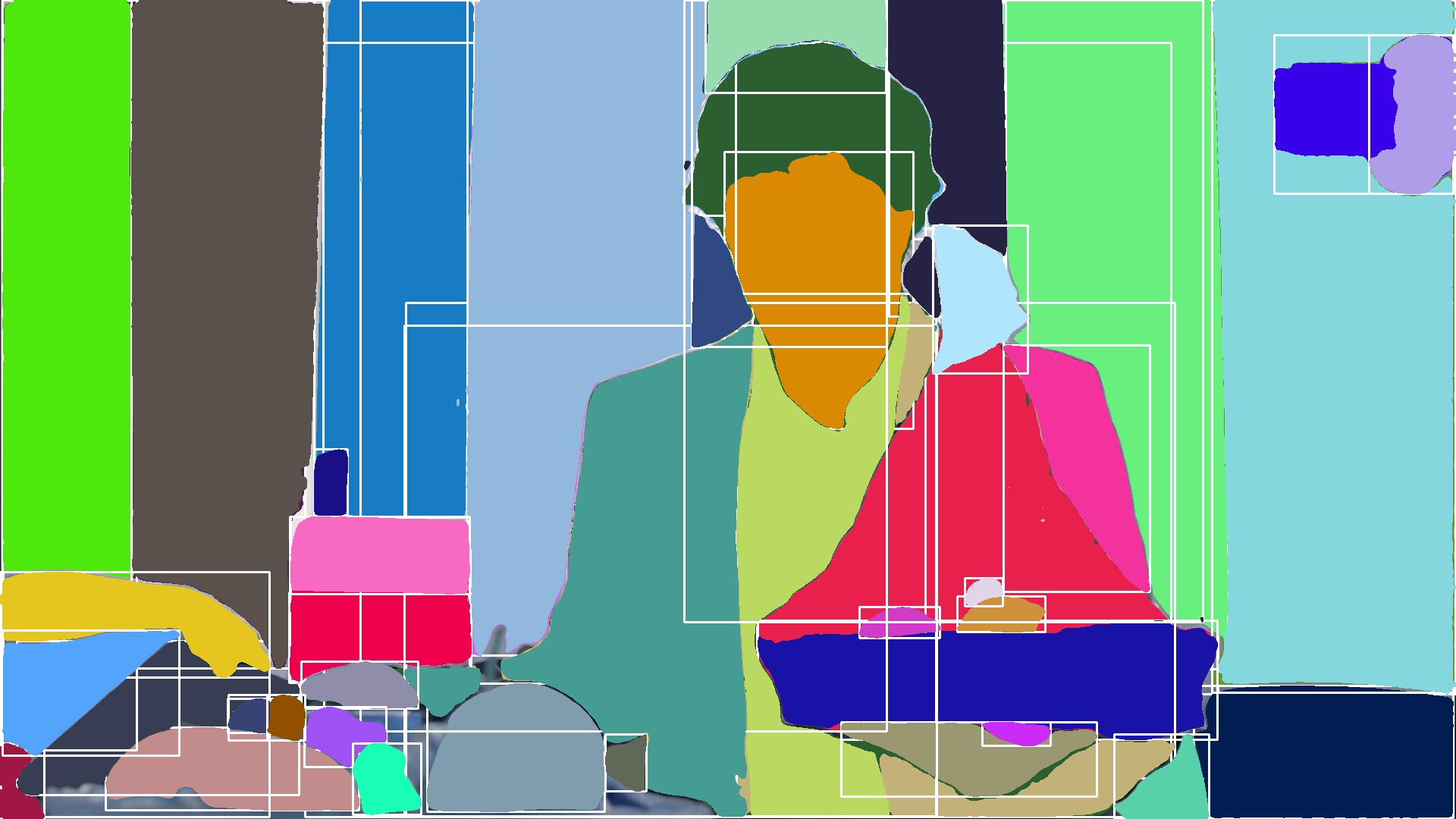}
        (b) SAM masks.
    \end{minipage} \\
    \begin{minipage}{.45\linewidth}
        \centering
        \includegraphics[width=\linewidth]{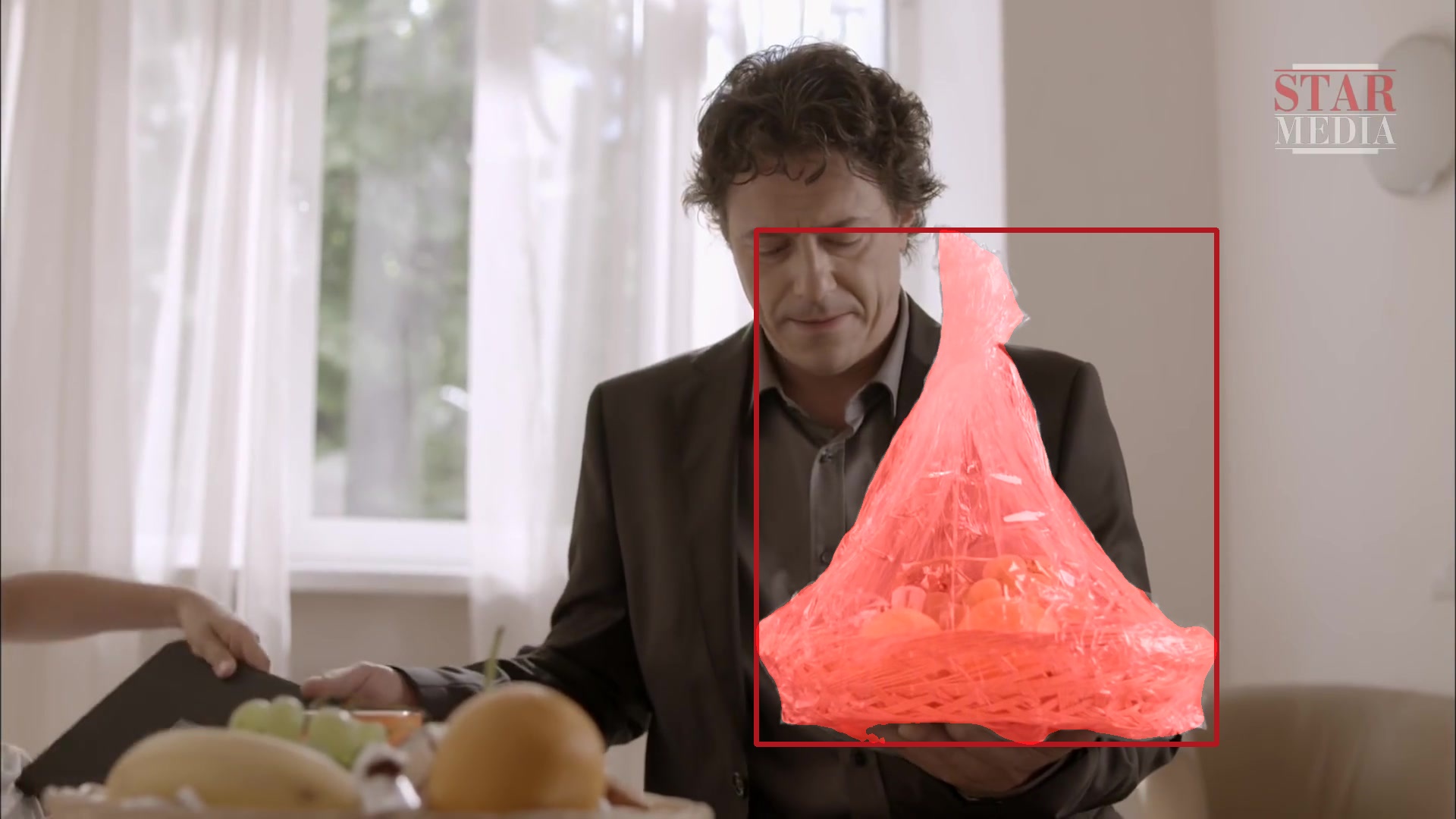}
        (c) Accurate bbox.
    \end{minipage}
    \begin{minipage}{.45\linewidth}
        \centering
        \includegraphics[width=\linewidth]{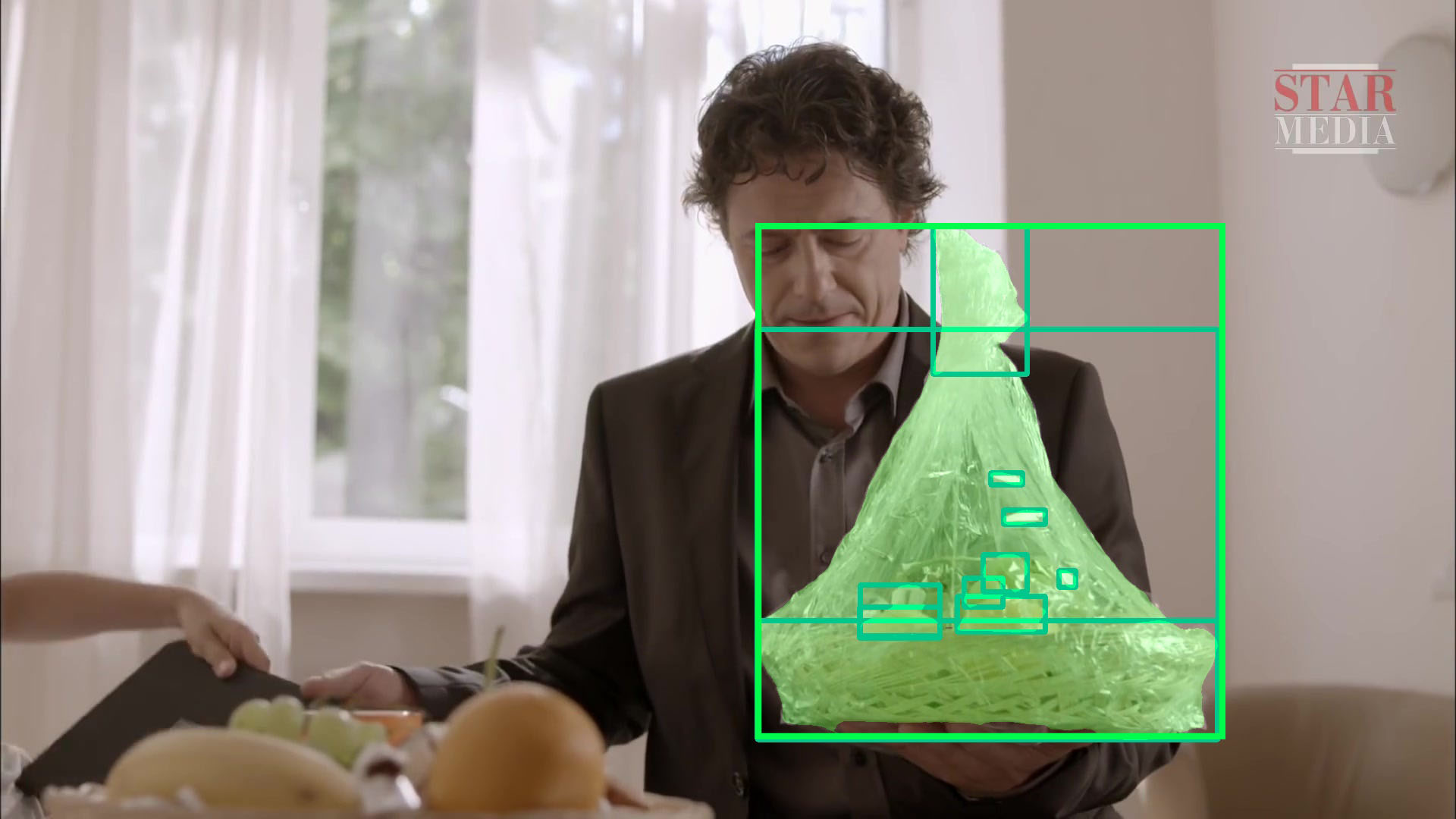}
        (d) GT mask\&bbox.
    \end{minipage}
    \vspace{-5px}
    \caption{SAM-based candidate generation.} 
    \vspace{-15px}
\label{fig:sam}
\end{figure}
Next, we calculate the area of the intersection between the \textit{proposal} masks and the \textit{accurate} mask ($A_{inter}$), and divide them by the area of the proposal masks $A_p$ to get a ratio for each proposal mask as $ratio=A_{inter}/A_o$.
Masks with a ratio greater than 0.9 are identified as GT masks.
Fig.~\ref{fig:sam} demonstrates the above process.

\subsection{Multi-Modal Feature}
\label{sec:MM}
To fully leverage the temporal and spatial continuity features of videos, including object information, HOI details, and spatial relations from multiple views, we employed a multi-modal feature extraction approach. 

{\bf Context Feature.} 
We utilize widely-used SlowFast~\cite{feichtenhofer2019slowfast} to extract context features from the video clip $C$.
The features from slow and fast branches are pooled along the time axis, then concatenated into the context feature map $f_c\in \mathcal{R}^{H\times W \times D}$, with $H\times W$ being the feature map resolution, and $D$ is the feature dim.

{\bf Object Feature.}
We first resize the masks $M_o^i$ of the $i$-th mask concerning the context feature map, then the feature for the $i$-th mask could be computed as 
\begin{equation}
    f_o^i = AvgPool(f_c \otimes M_o^i) \in \mathcal{R}^D,
\end{equation}
where $\otimes$ indicates element-wise multiplication. 
The object feature is denoted as $f_o\in \mathcal{R}^{N_o\times D}$ with $N_o$ masks.

{\bf Language Interaction Feature} is optional. 
If adopted, we input the language-guided query embedding $f_v\in \mathcal{R}^{D}$ of GroundingDINO~\cite{groundingdino}, which needs a language prompt and the key frame as input. Some other interaction features are discussed in Sec.~\ref{sec:ablation}.

{\bf 4D Human-Object Feature.}
Inspired by \citet{djrn}, which utilizes 3D information for HOI learning, we incorporate 3D information into our pipeline to exploit the rich HOI prior carried by 4D information.
Specifically, we lift GIO to 4D by reconstructing the HOIs in 3D.
However, lifting GIO to 4D is challenging given its diverse objects.
Existing efforts usually require 3D templates for the objects, which is inapplicable for open-world GIO.
To alleviate this, we adopt depth estimation for holistic scene estimation, bypassing the need for object templates.
Then, we align the human and scene for consistent 4D H-O representation.
Finally, we extract the 3D feature with the lightweight base point set (BPS)~\cite{bps}.

1) \textbf{Human reconstruction.}
Considering that videos \textit{without scene switching} allow for better human tracking and less processing time of 3D data, we first perform shot detection and segment the original video into multiple sub-clips. 
Then, PHALP~\cite{phalp} is adopted to recover 4D human tracklets from the sub-clips in SMPL~\cite{SMPL:2015} representation.
The 3D humans are further represented as SMPL mesh point clouds $p_h\in \mathcal{R}^{T \times N_h \times V \times 3}$, where $T$ is the length of the clip, $N_h$ is the number of existing human instances, and $V$ is the number of mesh vertices.

2) \textbf{Scene reconstruction via depth estimation}. 
We use ZoeDepth~\cite{zoedepth} to estimate the depth of the corresponding clip and transform them into scene point cloud $p_s\in \mathcal{R}^{T \times N_p \times 3}$, where $N_p$ is the number of points.

3) \textbf{Human-Scene alignment}.
The humans and scenes are initially inconsistent in scale and position. 
To align them, we render the $N_f$ front surface vertices $p_h^f \in \mathcal{R}^{N_f \times 3}$ of the human mesh to the image space, find the corresponding pixel of each vertice, and locate the corresponding point in the scene point cloud $p_s^f \in \mathcal{R}^{N_f \times 3}$.
Next, we align $p_h^f$ and $p_s^f$ by calculating the scale and displacement of $p_s^f$ to align with $p_h^f$. 
We calculate scale $s$ and displacement $b$ as
\begin{equation}
\begin{aligned}
d_h &= \frac{1}{{N_f}^2} \sum_{i=1}^{N_f} \sum_{j=1}^{N_f} \| {p_h^f}_i - {p_h^f}_j \|_2, \quad\quad \\
d_s &= \frac{1}{{N_f}^2} \sum_{i=1}^{N_f} \sum_{j=1}^{N_f} \| {p_s^f}_i - {p_s^f}_j \|_2, \\
s &= \frac{d_h}{d_s},{p_s^f}*=s, b = \frac{1}{N_f}\sum_{i=1}^{N_f} {p_h^f}_i - \frac{1}{N_f} \sum_{j=1}^{N_f} {p_s^f}_j.
\end{aligned}
\end{equation}
In detail, the scale is calculated as the ratio between the average pairwise distance of $p_h^f$ and $p_s^f$, while the displacement is calculated as the displace between the center point of $p_h^f$ and $p_s^f$.
The aligned human-scene point cloud is then formulated as $p=(p_h, p_s \cdot s + b) \in \mathcal{R}^{T\times (N_h \times V + N_h\times N_p) \times 3}$.

4) \textbf{3D feature extraction.}
We adopt BPS to extract features, which is simple and efficient for encoding 3D point clouds into fixed-length representations.
We randomly select $\frac{D}{2}$ fixed points in a sphere and compute vectors from these basis points to the nearest points in a point cloud; then use these vectors (or simply their norms) as features, shown in Fig.~\ref{fig:HPN}.
We adopt the human pelvis joint as the sphere center for base point generation.
We selected a radius of 1.5 times the height of the human body to cover the range of human interactions.
In this way, in one space, we obtain $T \times N_h \times \frac{D}{2}$ base points. 
We calculate the distances from these base points to the human mesh point cloud and the scene point cloud, treating them as features. 
Then we concatenate human features and scene features to get the final 3D feature $f_{3D}\in \mathcal{R}^{(T\times N_h) \times D}$, in the following we refer to $T\times N_h$ by $N_{3D}$, i.e., $\mathcal{R}^{N_{3D}\times D}$.

\subsection{Object Grounding}
\label{sec:decoder}
We utilize a 2D transformer decoder and a 3D transformer decoder to integrate multi-modal features. 
The 2D decoder outcome is sent to the 3D decoder as a query via an MLP as the 3D adapter. Note that the 2D decoder results have already been satisfactory, but the 3D decoder could further enhance predictions from the 3D perspective.
Each 2D decoder query $Q_s \in \mathcal{R}^{N_q\times D}$, is obtained via $Q_s=Q_v+Q_h$, where $Q_v \in \mathcal{R}^{N_q\times D}$ is the optional verb semantic query from the feature vector $f_v$, and the human query $Q_h \in \mathcal{R}^{N_q\times D}$ is obtained via a temporal pooling, a ROIAlign pooling, and a spatial pooling of the SlowFast features with the human bounding box.  
Given the context feature $f_c$, the object feature $f_o$, we concatenate them
as the key and value of the 2D decoder. The object feature $f_o$ and the 3D feature $f_{3D}$ are concatenated as the key and value of the 3D decoder.

The 2D/3D decoder outputs feature $f_q$.
The cosine similarity between $f_q$ and all object mask features $f_o$ is computed.
Then, we derive scores for each query relative to each mask, denoted as $S_m^i$. 
Higher scores suggest a greater likelihood of the mask being associated with the target object.
Considering that a person tends to interact with objects that are closer in proximity, we use the distance between masks and humans to assist us in calculating mask scores. 
The distance of the $i$-th mask is computed as $S_d^i = dist(C_h, C_m^i)$, where $C_h$ and $C_m^i$ refer to the human box and the $i$-th mask's box. Ultimately we adopt the GIoU~\cite{giou} distance.
The final score of the $i$-th mask is computed as
\begin{equation}
    \label{eq:dis}
    S_f^i = \gamma \times S_m^i + (1-\gamma) \times S_d^i,
\end{equation}
where $\gamma$ is a weight.
Then, we introduce a threshold $\tau$ to determine whether a mask is considered part of the target object. 
In the results for a certain query, if none of the mask scores exceed this threshold, we select the mask with the highest score. We cluster the predicted masks based on their depths and then determine the boundaries(detailed in the supplementary material).
For a given object w.r.t. $i$-th mask, BCE loss $L_o^i$ is used for supervision.
The overall loss is computed as $L_o = ({\textstyle \sum_{i=1}^{N_o} L_o^{i}} )/N_o $. 

\section{Experiments}
\label{sec:experiment}

\subsection{Setting}
\label{sec:task-metric}
Modified versions of mean Average Precision (mAP) and mean Intersection over Union (mIoU) are adopted.
For each GT tracklet, we sort all predictions by their scores in descending order. 
We identify the first prediction with an IoU higher than a threshold as a hit and calculate its precision by its position in that order.
mAP is averaged across all test instances. 
For mIoU, we calculate all IoUs between the GT and predicted boxes and report the largest IoU. 
To take into account the precision of the prediction, a \textit{weighted} mIoU is proposed as $\rm{mIoU_w}$.
For each GT tracklet, predictions are sorted by scores in descending order. 
The rank of each prediction is used to calculate $\rm{mIoU_w}$ as
\begin{equation}
\begin{aligned}
w(\hat{T_o}) &=\frac{1}{rank(\hat{T_o})}, \quad \\
mIoU_w(T_o)  &= \frac{\sum_{\hat{T_o}}w(\hat{T_o})IoU(T_o,\hat{T_o})}{\sum_{\hat{T_o}}w(\hat{T_o})},
\end{aligned}
\end{equation} \\
where $\hat{T_o}$ and $T_o$ denotes predicted and GT tracklets. 
Since $\rm{mIoU_w}$ is a more reasonable metric, we adopt $\rm{mIoU_w}$ instead of mIoU in the experiments.

\subsection{Implementation Details}
\label{sec:implementation}

For the 3D feature, considering the reconstruction quality of the 4D HOI layout, the reconstruction is only conducted for frames with object labels.
After filtering, there are 107,663 of 126,700 key-frames attached with 4D HOI layout (85,370 for training, 22,293 for inference).
SlowFast pre-trained on AVA 2.2 is adopted for video feature extraction.
An Adam optimizer, an initial learning rate of 1e-3, a cosine learning rate schedule, and a batch size of 16 are adopted. When training the 2D decoder, the learning rate of the parameters of SlowFast and Grounding DINO is 1e-5 and the 3D decoder is omitted. 
When training the 3D decoder, other parts except the 3D decoder are frozen.
$N_{3D}$ is set to 256, $N_o$ is set to 256 and $N_q$ is set to 24 for alignment.
Considering that the ground truth mask for each keyframe is sparse, we use weighted BCE loss, where the loss coefficient for true positions is ten times that of false positions. 

Our dataset supports different settings for further investigation, like inputting the interaction semantics to the grounding model, using more advanced LLM-extracted features, and the effect on the grounding of different human trackers, etc.
However, to focus on evaluating the grounding itself, we mainly discuss the default setting given the interaction semantics and GT human tracklets. 
For example, our system still predicts the interacted object well, without inputting the language interaction feature or the detected actions and humans from standard SOTA action detectors~\cite{feichtenhofer2019slowfast,videomaev2}. 
The proposed 4D-QA model has 246M parameters and achieves an inference speed of 8.63 FPS with a batch size of 1 (8 adjacent frames and 1 keyframe) on a single NVIDIA 3090 GPU.

\subsection{Baselines} 
\label{sec:baseline}
We adopt six models of four different types as our baseline. 
It is worth mentioning that since our task is new, we find these models most close to our task. But, they still do not fit our task very well in the setting. We devise corresponding protocols to adapt these models to our task.

\begin{table}[!t]
    \centering
    \caption{Results on GIO with multiple baselines.} 
    \vspace{-10px}
    \resizebox{\linewidth}{!}{
    \setlength{\tabcolsep}{3.5pt}
    \begin{tabular}{lcccccc}
        \hline
        \multirow{2}{*}{Methods}             & \multicolumn{5}{c}{mAPs}     &\multirow{2}{*}{mIoU$_w$}  \\
                                             & @0.5  & @0.6  & @0.7 & @0.8 & @0.9 &    \\ \hline
        PViC~\cite{zhang2023pvic}            & 11.78 & 9.89  & 7.73 & 5.31 & 2.45& 11.64          \\ 
        Gaze~\cite{gaze} & 12.71 &8.18 &5.43&3.14&1.17&16.06 \\
        Detic~\cite{detic}                   & 12.17  & 7.63  & 4.89 & 2.85 & 1.10 &  13.91       \\ 
        CG-STVG~\cite{cgstvg} & 12.35 & 8.97 & 6.17 & 3.70 & 1.72 & 17.50 \\
        Grounding DINO~\cite{groundingdino}  & 17.53 & 12.86 & 9.43 & 6.15 & 2.73  & 20.41           \\ 
        Qwen-VL~\cite{bai2023qwenvl} & 14.12 & 8.83 & 5.39 &  2.91 & 1.11  &20.28 \\ 
        \hline
        4D-QA & \textbf{23.38} & \textbf{18.48} & \textbf{14.40} & \textbf{10.71} & \textbf{6.39} & \textbf{29.71} \\ \hline
    \end{tabular}}
    \label{tab:res}
    \vspace{-15px}
\end{table}

{\bf Image/Video-based HOI models.} 
PViC~\cite{zhang2023pvic} and Gaze~\cite{gaze} are adopted as conventional image/video-based HOI detection baselines.
Given a frame or clip and a human bounding box $b$, the HOI models input the frame or clip and output a series of HOI triplets as $\langle b_h, b_o, p \rangle$, where $b_h, b_o$ are human and object bounding boxes, and $p$ is the predicted interaction probability.
We preserve all the results with $IoU(b, b_h)>0.5,p>0.2$, and the corresponding $b_o$ are adopted as the grounded objects.

{\bf Open-vocabulary object detection models.}
Detic~\cite{detic} is adopted, inputting a frame and expected object categories, outputting $\langle b_o, p \rangle$ as object bounding boxes and objectness score.
Results with $p>0.5$ are preserved and paired with the human query as the grounded objects.

{\bf Visual grounding models.}
Grounding DINO~\cite{groundingdino} 
is adopted, which takes a frame and a text prompt $s$ as input and produces $\langle b_o, p \rangle$ as grounded box and confidence. We also test a video-grounding baseline CG-STVG~\cite{cgstvg}, which aims to predict a spatial-temporal tube for a specific target subject/object given some semantic $s$.
$s$ is in the format as “The object that the person is \{interacting with\}”, where the placeholder ``\{Interacting with\}'' could be replaced with a specific action name. 
All the outputs are paired with the human query as the grounded object.

{\bf LLM based models.}
Qwen-VL~\cite{bai2023qwenvl} is adopted. It takes a frame and the text prompt ``Output the bounding box that the person is \{interacting with\}.'' and produces the bounding box $b_o$ if detected. 

\subsection{Results}
\label{sec:result}
Results are shown in Tab.~\ref{tab:res}.
For all the models, we combine the human-object distance for $\rm{mIoU_w}$ and mAP as Eq.~\ref{eq:dis}.
All baselines provide sub-optimal performance, indicating their deficiency in interactiveness grounding.
Also, most baselines take little use of temporal information since they utilize only images as inputs, leading to bad performances on ``temporally hidden objects'' such as the chairs obstructed by humans sitting on them but appearing in the next frame. 

As HOI detection models, PViC and Gaze fail to perform well due to the large number of novel objects in GIO. 
The open-vocabulary object detection model Detic demonstrates low $mIoU_w$ since it cannot discriminate the interacted objects related to humans (interactiveness~\cite{tin}).
It is noteworthy that Detic tends to predict a substantial number of object bounding boxes, sometimes exceeding 900, with many false positive predictions.
CG-STVG, lacking pre-training on large datasets and integration of visual-language models, outperforms PViC, Gaze, and Detic in mIoU$_w$, using a single high-quality bounding box per HOI instance for higher mIoU$_w$ despite lower mAP.
Grounding DINO performs better than other baselines, but it is still limited for ``hidden objects''. 
Also, it frequently fails to fully understand the interaction semantics. 
Qwen-VL, a large vision language model, provides decent $mIoU_w$ but poor $mAP$s, which suggests that although Qwen-VL can localize the approximate positions of most objects, it struggles to detect precise bounding boxes.
Our model performs well in localizing diverse and unseen objects, where the baselines struggle.
Also, our model demonstrated decent $mIoU_w$. 
These experimental findings indicate that our method excels in object grounding for spatiotemporal HOI understanding.

In addition, we considered ST-HOI as the task design, resulting in the highest mAP of \textbf{6.8},
i.e., the ST-HOI task is kind of \textbf{too challenging} even ignoring the annotation missing problem. This demonstrates the rationality and exploratory potential of the GIO task formulation.

We also evaluate 4D-QA on VidHOI~\cite{chiou2021st} with the GIO task. 
The GIO-pretrained 4D-QA gets 14.23 mAP and 22.66 mIoU$_w$ and \textbf{25.35} mAP and \textbf{29.61} mIoU$_w$ after 10 epoch finetuning. GroundingDINO gets \textbf{15.87} mAP and \textbf{17.57} mIoU$_w$.
Results on VidHOI reveal that \textit{grounding interacted objects} is challenging enough to be carefully explored and our 4D-QA maintains decent performance. 
VidHOI only has \textbf{1.22} HOIs/frame in the filtered (Sect.~\ref{sec:datastatics}) test set, allowing GroundingDINO to slightly outperform zero-shot 4D-QA on mAP because Grounding DINO performs better to localize the only object in the one-HOI frame.

\subsection{Visualization}
\begin{figure}[!t]
    \centering
    \includegraphics[width=\linewidth]{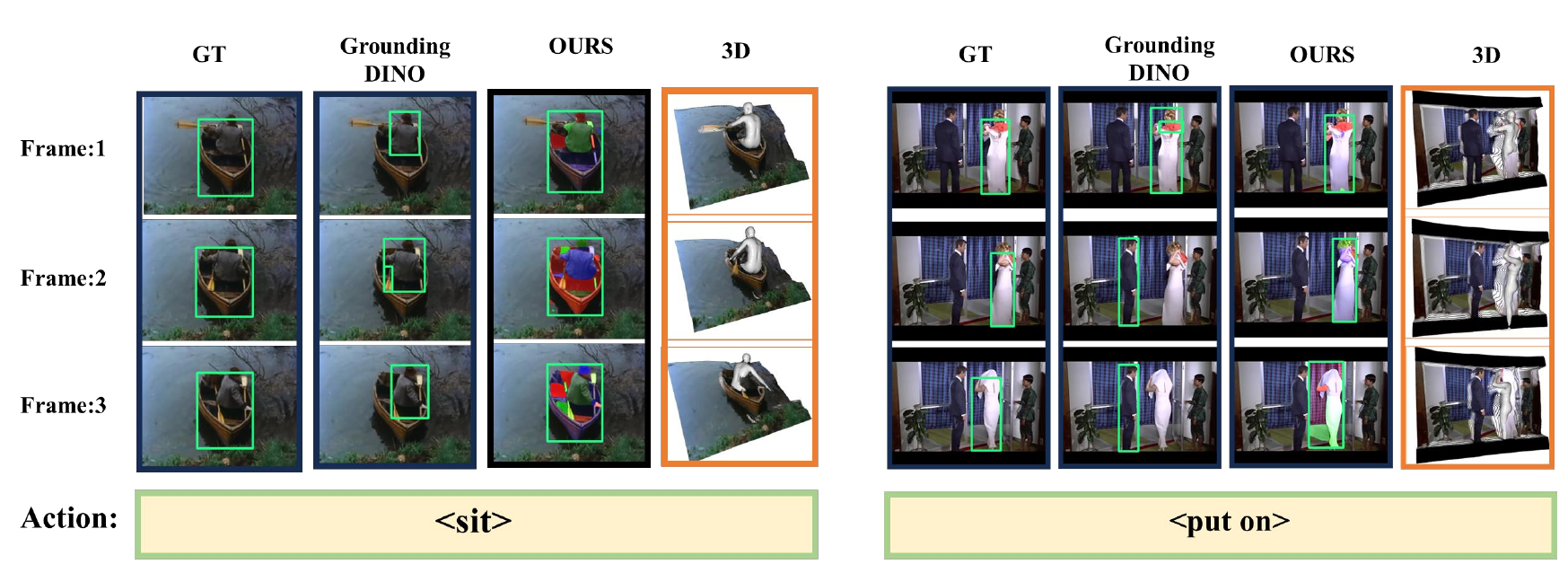}
    \vspace{-15px}
    \caption{Visualization of interacted object grounding. We also list the reconstructions.}
    \label{fig:visualization}
    \vspace{-10px}
\end{figure}

Fig.~\ref{fig:visualization} visualizes the grounded interacted object in 3 consecutive frames. The predicted masks (colored regions) are integrated into final object boxes (green) as in Sec.~\ref{sec:decoder}.

\begin{figure}[!t]
    \begin{minipage}{\linewidth}
        \centering
        \includegraphics[width=.85\linewidth]{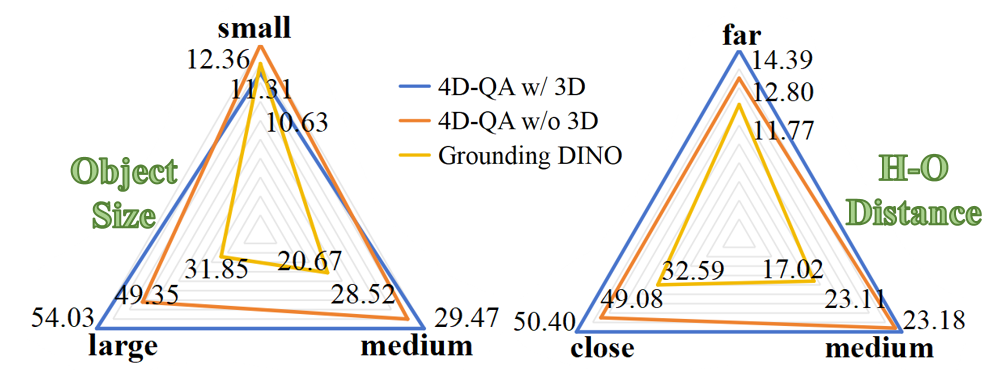}\\
        (a) mIoU$_w$ w.r.t. object size and H-O distance.
    \end{minipage} \\
    \begin{minipage}{\linewidth}
        \centering
        \includegraphics[width=.98\linewidth]{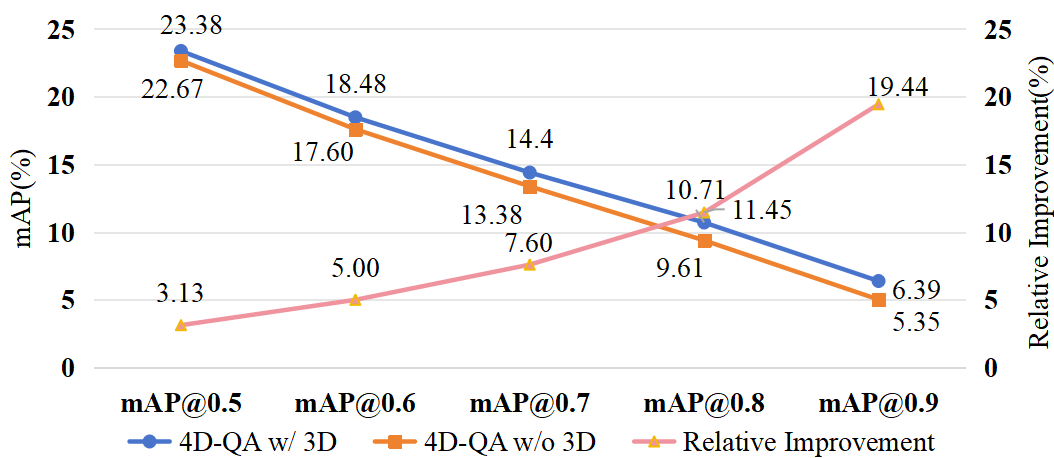}\\
        (b) mAP w.r.t. IoU threshold.
    \end{minipage}
    \vspace{-8px}
    \captionof{figure}{Fine-grained performance analysis.}
    \label{fig:compare}
    \vspace{-15px}
\end{figure}

\begin{table}[!t]
    \centering
    \caption{Ablation Results.} 
    \vspace{-5px}
    \resizebox{0.8\linewidth}{!} {
    \begin{tabular}{lcc} \toprule
        Model & mAP@0.5 & mIoU$_w$\\ \midrule
        4D-QA & \textbf{23.38} & \textbf{29.71} \\ \hline
        4D-QA w/o 3D & 22.67 & 28.76 \\ \hline
        4D-QA w/ $L_2$ distance & 22.34 & 28.39 \\
        4D-QA w/o distance & 22.07 & 27.85 \\ \hline
        4D-QA w/ Bert & 20.02 & 28.11 \\
        4D-QA w/ CLIP & 20.00 & 28.15 \\ 
        4D-QA w/o action & 19.04 & 27.30 \\ \hline
        4D-QA w/ pred hbox w/o action & 18.92 & 27.17 \\ \hline
        4D-QA w/ box regression & 18.82 & 26.88 \\ \bottomrule
    \end{tabular}}
    \vspace{-15px}
\label{tab:abla}
\end{table}

\subsection{Ablation Study}
\label{sec:ablation}
We conduct ablation studies on the different components of our model on the GIO test set as reported in Tab.~\ref{tab:abla}. 

{\bf Distance.}
Removing the use of distance would result in a degradation (\textbf{22.07} mAP, \textbf{27.85} mIoU$_w$). Replacing the GIoU distance with the L2 distance would also cause a decline in performance (\textbf{22.34} mAP, \textbf{28.39} mIoU$_w$). 

{\bf 3D Feature $f_{3D}$.}
4D-QA w/o $f_{3D}$ presents a performance decrease (\textbf{22.67} mAP, \textbf{28.76} mIoU$_w$). 
Fig.~\ref{fig:compare}(a) further shows the influence of $f_{3D}$ w.r.t. different data characteristics, namely the relative object size and H-O distance.
As shown, our methods provide superior performance compared to Grounding DINO across different data groups, especially on medium-to-large objects and close-to-medium H-O pairs. For 2D and 3D's difference, $f_{3D}$ gains \textbf{4.68} mIoU$_{w}$ on large objects and \textbf{1.32} mIoU$_{w}$/\textbf{1.59} mIoU$_{w}$ on close/far H-O pairs. 
Besides, we find $f_{3D}$ outperforms 2D by \textbf{2.25} mIoU$_w$ and \textbf{2.83} mAP@0.5 on \textbf{40} verb classes, especially on verbs like \texttt{drive, exit, press} that involve large or occluded objects or occur in complex scenarios.
Notably, in Fig.~\ref{fig:compare}(b) the relative improvement that $f_{3D}$ brings increases with the IoU threshold requirement for mAP, indicating that $f_{3D}$ contains more accurate predictions than 2D.

{\bf Interaction Feature.}
We replaced the Grounding DINO interaction feature with the Bert and CLIP interaction language embedding. 
In addition, we performed a test without the interaction feature. 
As shown, the sophisticated feature from the Grounding DINO can help the grounding task to better utilize the interaction information, while the simple language representation difference between Bert and CLIP affects the performance little.
Eliminating the interaction feature brings a major performance degradation.

{\bf Predicted human boxes}, with 88 mIoU w.r.t. GT human boxes, were used as human queries. The slight performance drop indicates the robustness and flexibility of 4D-QA.

{\bf Box Regression.} We used an MLP after the decoder to directly regress the boxes instead of utilizing SAM mask candidates or other box proposals. The performance drop shows the importance of SAM-generated mask candidates.

\section{Conclusion}
We constructed GIO, which consists of many rare objects that are overlooked but important in HOI learning.
290K frame-level HOI triplets annotations with 1,098 objects were collected.
Based on GIO, an interacted object grounding task was devised and a 4D-QA framework was proposed to tackle this challenging task with decent results.  
We believe GIO would inspire deeper activity understanding and interactive object grounding, thus enhancing the performance of tasks associated with spatiotemporal analysis and exploration.

\section*{Acknowledgement}
This work is supported in part by the National Natural Science Foundation of China under Grants No.62302296, 62306175, and 62472282.

%

\bibliography{main}

\appendix

\section*{Appendix Overview}
The contents of this supplementary material are:

Sec.~\ref{sec:characteristics}: Characteristics of GIO.

Sec.~\ref{sec:method-detail}: Method Details.

Sec.~\ref{sec:results}: More Experiments on GIO.

Sec.~\ref{sec:discussion}: Discussion.

\section{More Characteristics of GIO}
\label{sec:characteristics}

\subsection{Selecting Videos for GIO Test Set}
\label{sec:video-select}
To make GIO challenging and practical, we construct its test set by seeing video selection as a \textbf{multi-objective integer programming} problem. 
Note that further subdivision into validation and test sets is performed here and in the mentioned GIO test set below during the experiments. However, for the sake of uniformity, we will refer to them collectively as the test set in this context.

\textbf{First}, given the video number $N$, interaction class number $N_a$, object class number $N_o$ and GT object location heatmap size $N_h$ (the original size of AVA~\cite{AVA} frames is resized to the size of the heatmap, and here we use a 1D vector to represent the values of original 2D heatmap) in AVA train-val sets, we define a binary variable $x_{i}\in\{0, 1\}, 1\le i \le N$ for each video to indicate whether to choose it or not for our test set. We restrict the sum of $x_{i}$ to the number of videos in the test set ($N_{t}$) according to a certain split ratio.

\textbf{Second}, for video $i$, we calculate its distributions of interaction class $\mathbf{a_i}\in \mathbb{N}^{N_a}$ (a set of interaction class frequencies), object class $\mathbf{o_i}\in \mathbb{N}^{N_o}$ (a set of object class frequencies), and object location GT heatmap $\mathbf{c_i}\in \mathbb{N}^{N_h}$. 
Each $x_{i}$ is multiplied to $\mathbf{a_i}$, $\mathbf{o_i}$ and $\mathbf{c_i}$ individually, then we add them up respectively to obtain three total distributions $\sum_{i=1}^N\mathbf{a_i}x_i$, $\sum_{i=1}^N\mathbf{o_i}x_i$, and $\sum_{i=1}^N\mathbf{c_i}x_i$ for all videos.

\textbf{Finally}, we want the test set to contain as many as possible interactions, object classes, and diverse object locations to fully evaluate the models. 
To this end, we calculate the variances $Var\left(\sum\limits_{i=1}^N\mathbf{a_i}x_i\right)$ and $Var\left(\sum\limits_{i=1}^N\mathbf{o_i}x_i\right)$ of interaction and object class distributions, use the variances as minimization objectives to search the suitable videos with the highest varieties of interaction and object classes. 
Moreover, we find that many objects are located at the half bottom of frames. Thus, to increase the variety of object location, we restrict the distribution of the top half part of heatmaps $\sum\limits_{i=1}^N\sum\limits_{k=1}^{N_h/2}\mathbf{c_{i, k}}x_i$ to a given threshold $\gamma$.
Additionally, to preserve the frequencies of some interaction classes from degrading to zero, we also add external restrictions on $\mathbf{a_i}$ with a threshold $\alpha_j$ for each interaction class $j$.  
The final programming problem to be solved is
\begin{align*}
&\min\quad z=Var\left(\sum\limits_{i=1}^N\mathbf{a_i}x_i\right)+Var\left(\sum\limits_{i=1}^N\mathbf{o_i}x_i\right) \\
& \begin{array}{c@{\quad}c}
s.t.&x_i\in\left\{0, 1\right\}, \quad1\le i \le N\\
    &\sum\limits_{i=1}^Nx_i=N_t,\\
    &\sum\limits_{i=1}^N\mathbf{a_{i, j}}x_i\geq\alpha_j,\\
    &\sum\limits_{i=1}^N\sum\limits_{k=1}^{N_h/2}\mathbf{c_{i, k}}x_i\geq\gamma.\\
\end{array}
\end{align*}
At last, the results are used to select the videos for our test split.
 
\begin{figure}[!t]
\centering
    \begin{minipage}{.5\linewidth}
        \centering
        \includegraphics[width=\linewidth]
        {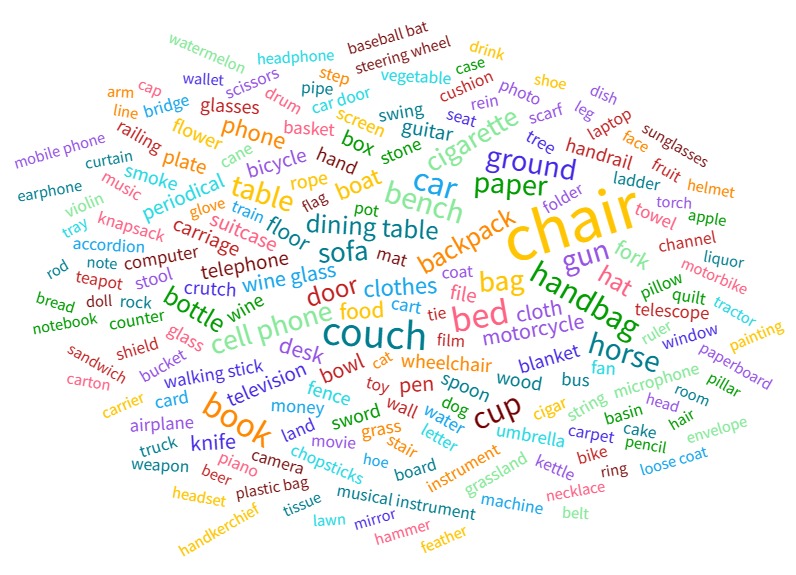}     
        (a) Object classes in GIO.
    \end{minipage}
    \begin{minipage}{.4\linewidth}
        \centering
        \includegraphics[width=\linewidth]{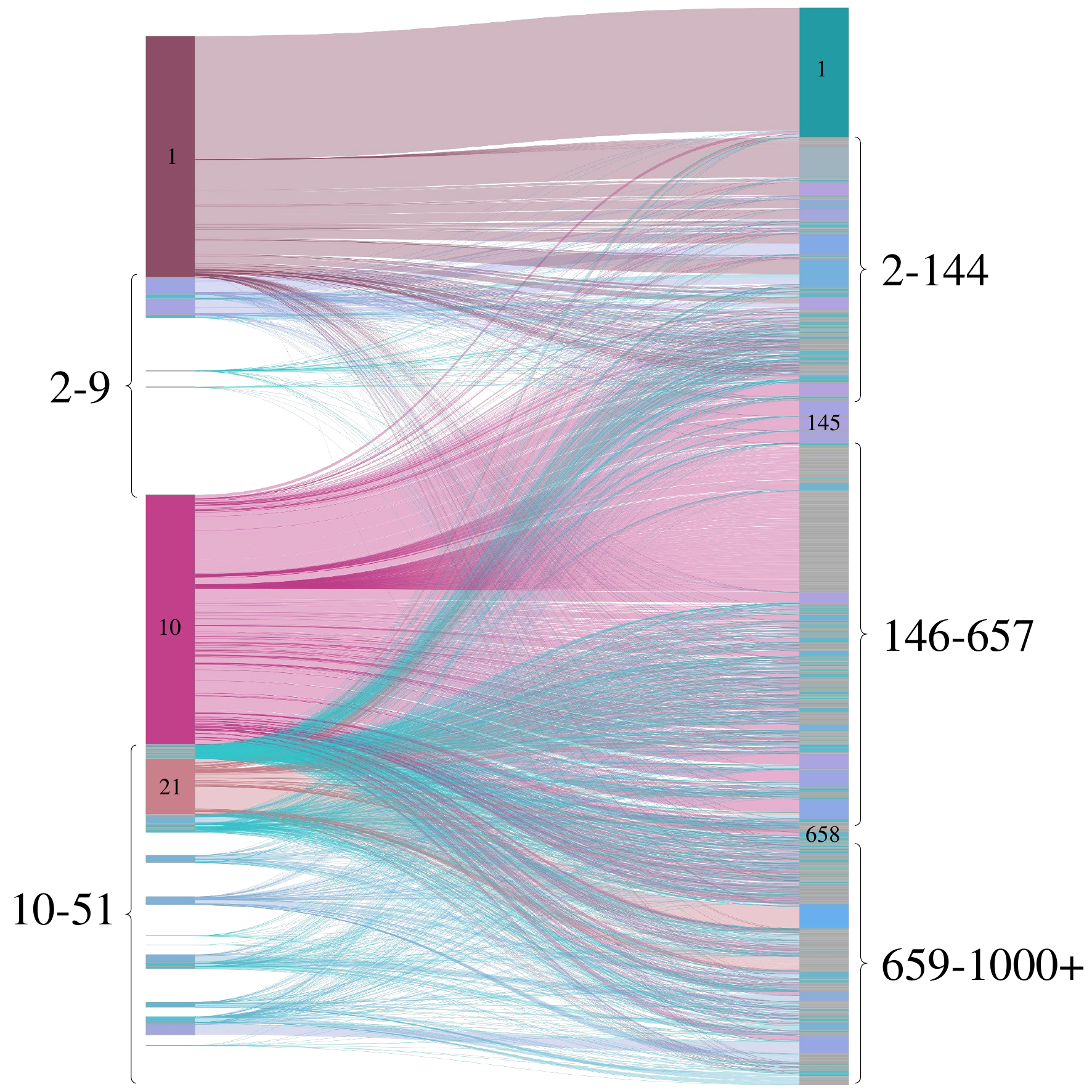}
        (b) Verb-object co-occurrence.
    \end{minipage}
    \caption{(a) shows the frequency of occurrence of object categories in GIO. The width of each 
line in (b) represents the ratio of one action-object interaction.}
    \label{fig:characteristics}
\end{figure}

The object classes in GIO and verb-object co-occurrences are demonstrated in Fig.~\ref{fig:characteristics}.

\subsection{Statistics of Data Split}

The detailed statistics of the data split are shown in Tab.~\ref{tab:dataset table}.

\begin{table}[!t]
\centering
\caption{Statistics of data split.}
\resizebox{.6\linewidth}{!}{
\begin{tabular}{c c c c}
\hline
Split & Box & Tracklet & Frame\\ \hline
Train & 234K & 104K & 102K\\
Test & 56K & 22K & 29K\\
All & 290K & 126K & 131K\\
\hline
\end{tabular}}
\label{tab:dataset table}
 \vspace{-10px}
\end{table}

\begin{table*}[t]
\begin{center}
\caption{Interaction class list of GIO.}
\label{tab:action list}
\resizebox{\linewidth}{!}{
\begin{tabular}{c|c|c|c|c|c|c|c}
\hline
Action Id&Action Class&Action Id&Action Class&Action Id&Action Class&Action Id&Action Class \\ 
\hline
1 & jump/leap & 14 & dress/put on clothing & 27 & paint&40 & shoot\\
2 & lie/sleep & 15 & drink & 28 & play board game&41 & shovel\\
3 & sit & 16 & drive & 29 & play musical instrument&42 & smoke\\
4 & answer phone & 17 & eat & 30 & play with pets&43 & stir\\
5 & brush teeth & 18 & enter & 31 & point to&44 & take a photo\\
6 & carry/hold & 19 & exit & 32 & press&45 & text on/look at a cellphone\\
7 & catch & 20 & extract & 33 & pull&46 & throw\\
8 & chop & 21 & fishing & 34 & push&47 & touch\\
9 & clink glass & 22 & hit & 35 & put down&48 & turn\\
10 & close & 23 & kick & 36 & read&49 & watch\\
11 & cook & 24 & lift/pick up & 37 & ride&50 & work on a computer\\
12 & cut & 25 & listen & 38 & row boat&51 & write\\
13 & dig & 26 & open & 39 & sail boat& & \\
\hline
\end{tabular}}
\end{center}
\vspace{-5px}
\end{table*}

\subsection{Data Samples}

\begin{figure*}
    \centering
    \includegraphics[width=0.95\linewidth]{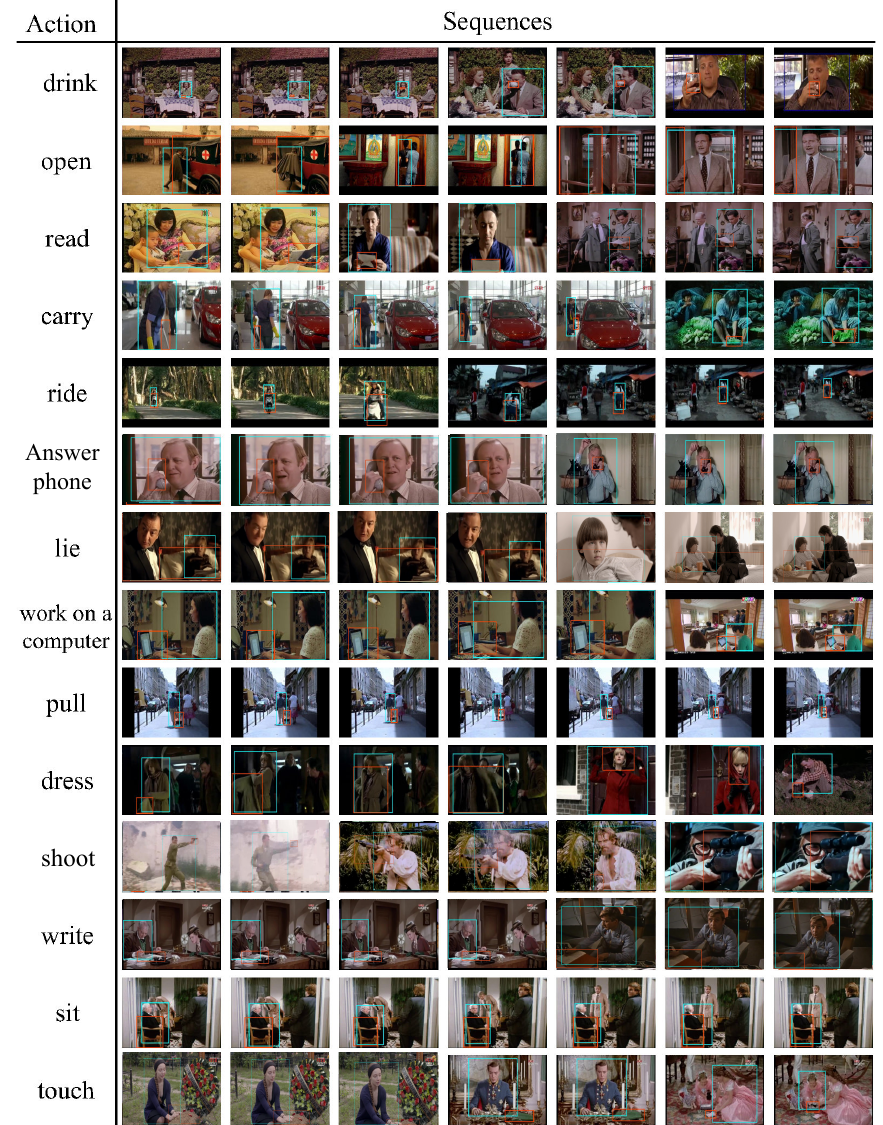}
    \caption{Data samples and their ST-HOI labels in GIO.}
    \label{fig:Data Sample}
\end{figure*}

Some data samples of GIO are shown in Fig.~\ref{fig:Data Sample}.
Human and object GT boxes are in blue and red respectively. 

\begin{algorithm}[t]
\SetAlgoLined
\KwIn{object class list $W$=$\{w_1, w_2, ...\}$}
\KwOut{cluster list $C$=$\{C_1, C_2, ...\}$}
 Initialize empty cluster list $C$\;
 \For{$i$=1:$|W|$}{
   \eIf{$|C|>0$}{
     \For{$j$=1:$|C|$}{
        Get $\hat{w}_j \in C_j$ with highest WordNet level\;
        \eIf {WordNet has path between ($w_i$, $\hat{w}_j$)}{
            Add $w_i$ to $C_j$\;
        }   
        {
            Add $w_i$ as a new cluster to $C$\;
        }
     }
   }{
    Add $w_i$ as a new cluster to $C$\;
  }
  $i$++\;
 }
 \caption{Clustering object classes.} 
 \label{alg:clustering}
\end{algorithm}
 \vspace{-5px}

\subsection{Interaction List}
The detailed interaction classes are listed in Tab.~\ref{tab:action list}.

\subsection{Object Class Taxonomy}
To deal with the diversity of object class annotations in GIO, following EPIC-Kitchens~\cite{epic-kitchen}, we use WordNet~\cite{wordnet} to construct an object class tree. The detailed procedure is as follows:
\begin{itemize}
\item First, with the annotated object class list $W$=$\{w_1, w_2, ...\}$, we follow the clustering procedure of Algorithm \ref{alg:clustering} to build a cluster list $C$.
\item Then, we find some object classes are wrongly clustered due to the polysemy.
For example, the first explanation of ``banana" in WordNet is a kind of ``herb", instead of ``fruit". For these classes, we manually remove them from $C$, correct their explanations, and add them to $C$ as unique clusters.
\item Finally, we follow Algorithm \ref{alg:class_tree} to construct the object class tree with the clusters from $C$ and correct the ambiguous class names.
\end{itemize}

\begin{algorithm}[t]
\SetAlgoLined
\KwIn{cluster list $C$=$\{C_1, C_2, ...\}$}
\KwOut{object class tree $T$}
\SetKw{Break}{break}
\SetKwFunction{FMain}{ConstructTree}
\SetKwProg{Fn}{Function}{:}{}
\Fn{\FMain{$C_i$}}{
        \tcp{Construct object class tree $T$ from cluster $C_i$.}
        Initialize $T$ from the first word $w_1$ of $C_i$\;
        \For{$j$=2:$|C_i|$}{
            Get the $j$-th word $w_{i,j}$ of cluster $C_i$\;
            Get the node $T_k \in T$ with the shortest path between ($w_{i,j}$, $T_k$) in WordNet\;
            Add $w_{i,j}$ to $T_k$\;
         }
        \textbf{return} $T$
}
\SetKwFunction{FMain}{CombineTree}
\SetKwProg{Fn}{Function}{:}{}
\Fn{\FMain{$T_x$, $T_y$}}{
        \tcp{Combine object class tree $T_x$ and $T_y$.}
        Find root nodes $R_x$ and $R_y$ of $T_x$ and $T_y$\;
        Find closest common parent $R_{xy}$ of $R_x$ and $R_y$ in WordNet\;
        Add $R_x$ and $R_y$ to the children of $R_{xy}$\;
        Construct new class tree $T_{xy}$ from $R_{xy}$\;
        \textbf{return} $T_{xy}$
}
 Initialize object class tree $T$=ConstructTree($C_1$)\;
 \For{$i$=2:$|C|$}{
   $T_i$ = ConstructTree($C_i$)\;
   $T$ = CombineTree($T$, $T_i$)\;
 }
 \caption{Constructing object class tree.} 
 
 \label{alg:class_tree}
\end{algorithm}
The detailed object classes are listed in the \textbf{GIO-object-classes.csv}.
The detailed object class tree according to WordNet~\cite{wordnet} can be found in our \textbf{GIO-object-class-tree.html} file.

\subsection{Statistics of Action, Object, and Tracklet Length}
\begin{figure}[t]
    \centering
    \includegraphics[width=\linewidth]{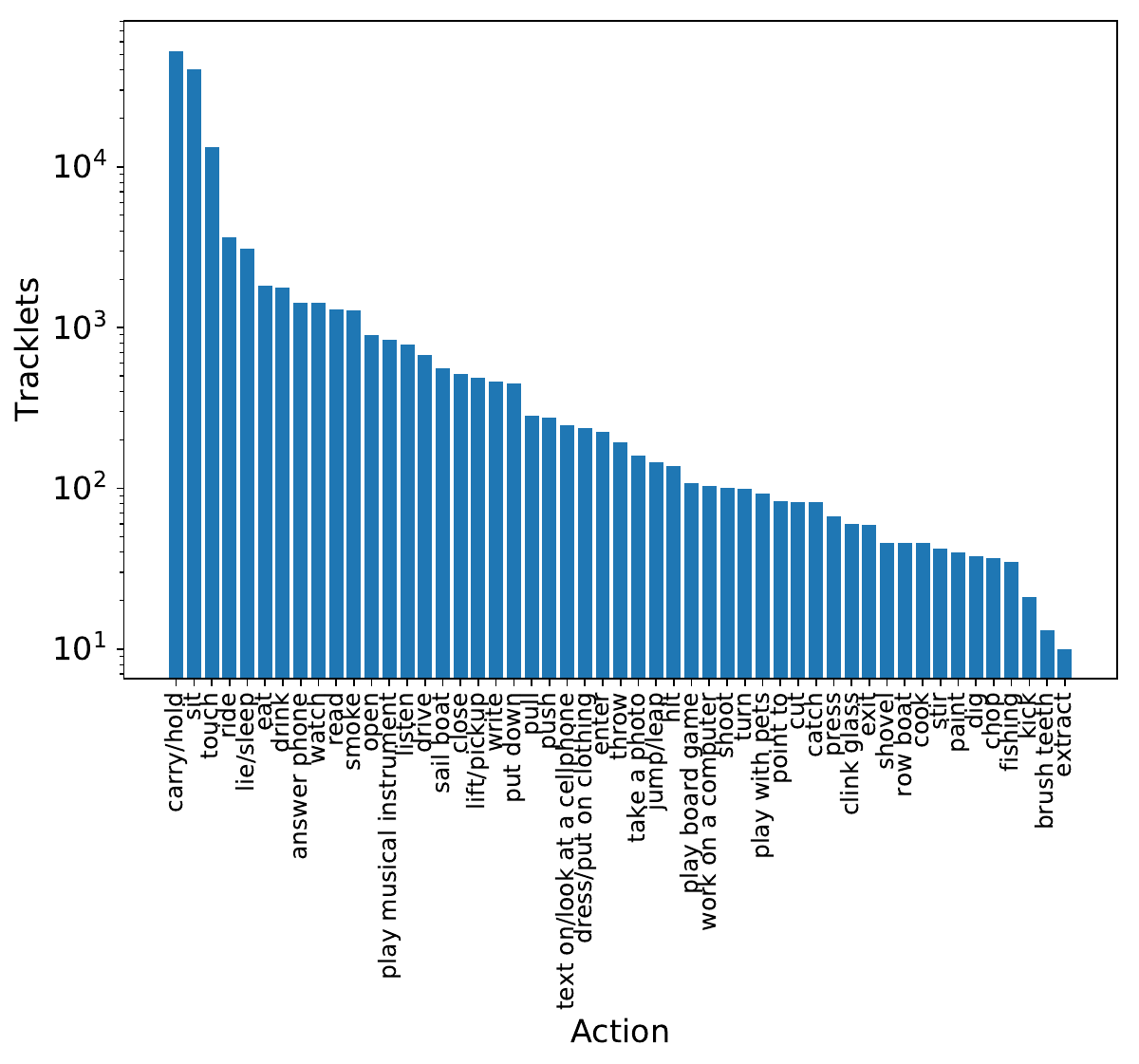}
    \caption{The distribution of tracklet number per action.}
    \label{fig:action}
\end{figure}

\begin{figure}[t]
    \centering
    \includegraphics[width=\linewidth]{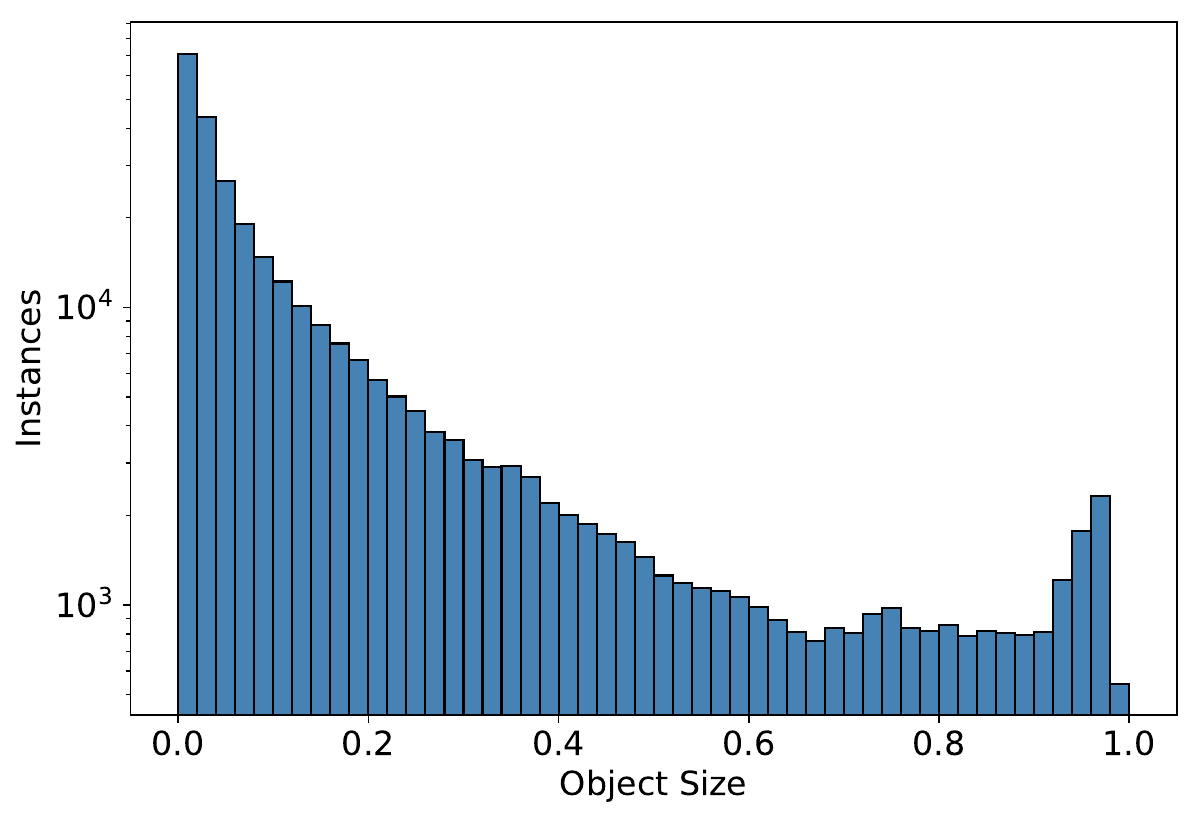}
    \caption{The distribution of normalized object size.}
    \label{fig:object}
\end{figure}

\begin{figure}[t]
    \centering
    \includegraphics[width=\linewidth]{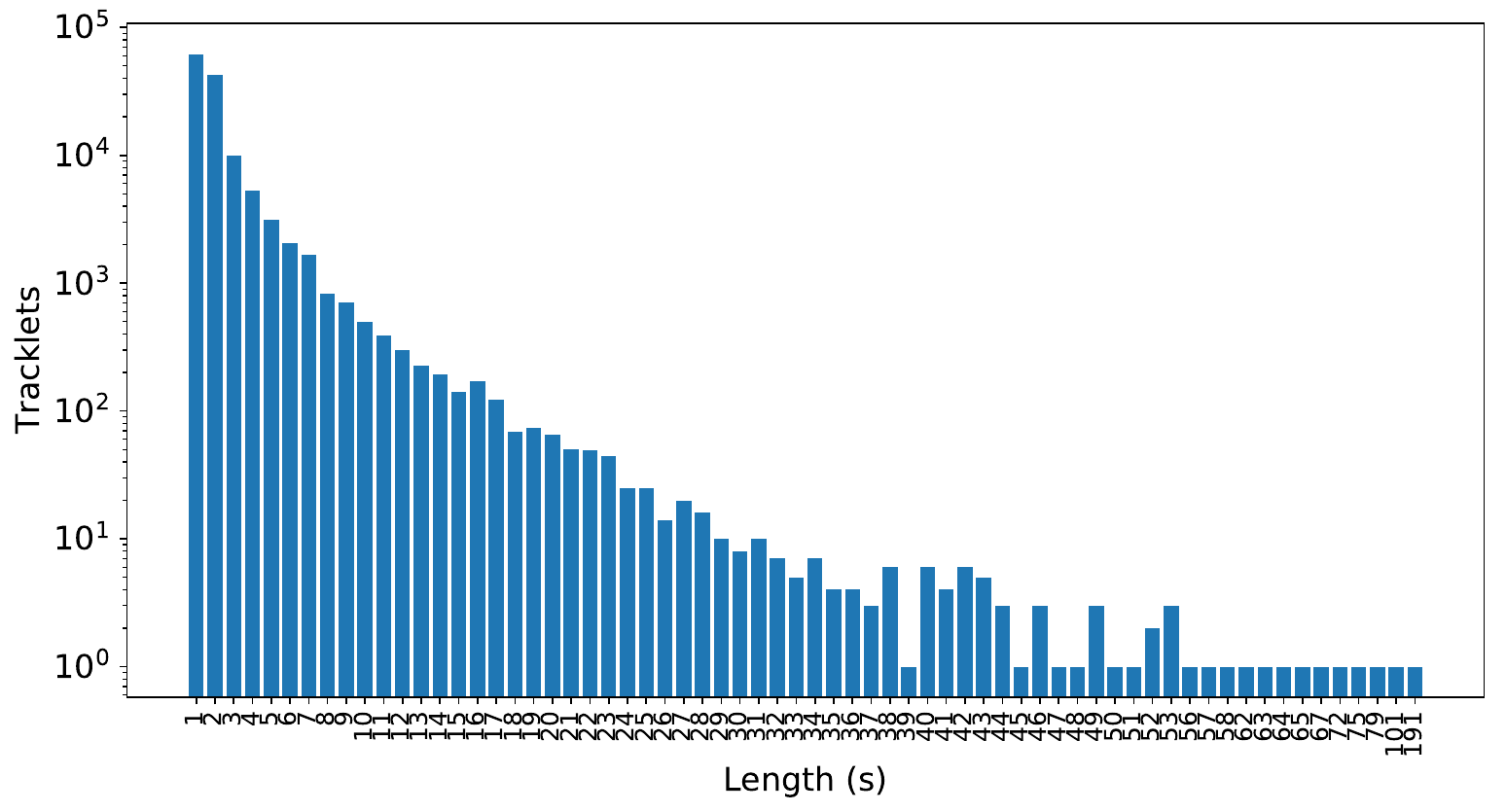}
    \caption{The distribution of tracklet length.}
    \label{fig:length}
\end{figure}

We also provide the distribution of action, object, and tracklet length of GIO in Fig.~\ref{fig:action}-\ref{fig:length}.


\section{Method Details}
\label{sec:method-detail}
\begin{figure}
  \centering
  \includegraphics[width=\linewidth]{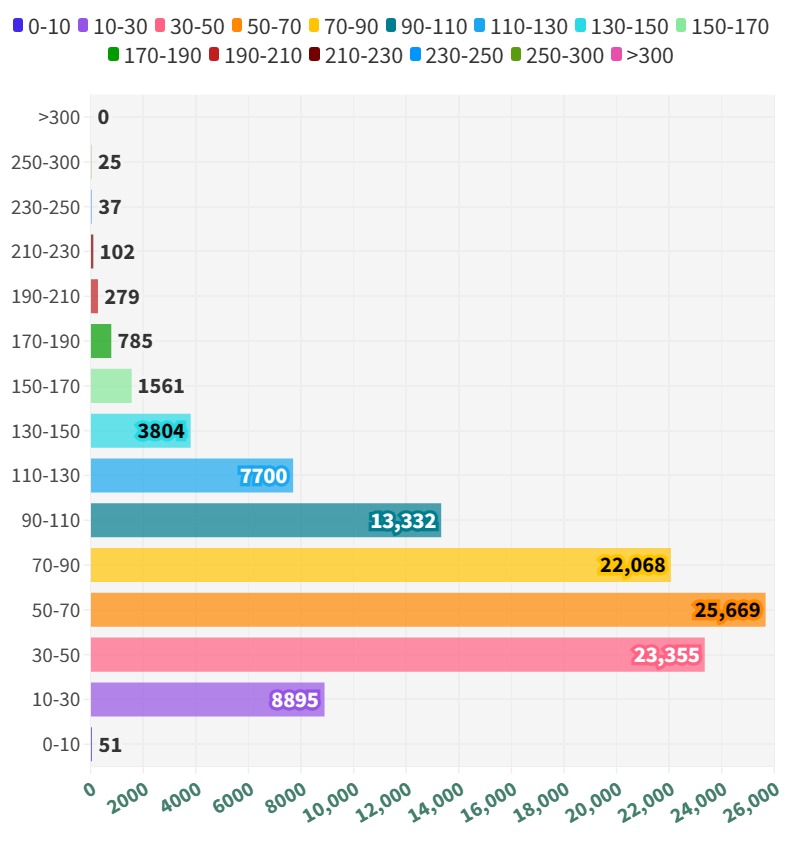} 
  \caption{Distribution of mask numbers in keyframes.}
  \label{fig:mask_dis}
\end{figure}

\vspace{5pt}
\textbf{SAM-based Candidate Generation.}
SAM is first fed with a grid of point prompts over the image.
Then, low-quality and duplicate masks are filtered out. 
As a result, each image would produce at most 255 masks. 
The distribution of mask numbers is shown in Fig.~\ref{fig:mask_dis}. 
It can be seen that the number of masks in most images is distributed in the range of $30 \sim 110$.
The masks roughly follow a normal distribution on key frames of the dataset. This ensures a certain level of robustness in the training of our model.

{\bf Strategy of Bounding Box Generation.} 
For 4D-QA, we get all the masks together with the scores. We use the threshold $\tau$ to determine there are $p$ masks considered as part of the target object, denoted as $\{M_t\}^{p}_{t=1}$. We will get two boxes from $\{M_t\}^{p}_{t=1}$. The one box is the minimum bounding box that can contain all $\{M_t\}^{p}_{t=1}$. For the other, we select the mask with the highest score, denoted as $m_h$, and also introduce a threshold $\beta$, to filter the $\{M_t\}^{p}_{t=1}$ by calculating the depth difference. The masks with a difference of less than $\beta$ are considered to belong to the same cluster as $m_h$. Then we get the selected masks together with $m_h$ forming the second bounding box.
For 4D-QA w/o 3D, depth information should not be utilized in this process. So only the first bounding box is adopted. 

{\bf Ablation Study Details.}
The relative object size is calculated as $r_{o|h}=a_o/a_h$, with human area $a_h$ and object area $a_o$.
Then, we divide the test set into three splits: small($r_{o|h} \le 0.3$), medium($0.3 < r_{o|h} \le1.0$), large($r_{o|h} > 1$). 
H-O distance is indicated by $r = GIoU(hbox, obox)$.
And we divide them into three splits: far($r\le0.04$), medium ($0.04<r\le0.22$), and close($r>0.22$). 

{\bf Other Details.}
We used 4 NVIDIA GeForce RTX 3090 GPUs, each equipped with 24GB of memory. The depth of a mask is determined by the mode of the depth values of each pixel. Extensive experiments have demonstrated that the mode can encompass the most pixels within the same neighboring interval. There are three important hyper-parameters for the mask post-processing, i.e., $\gamma, \tau, \beta$. We use grid search to systematically explore different combinations of these hyper-parameters and identify the optimal values that maximize the model's performance. The parameters will be public together will our code.

\section{More Experiments on GIO}
\label{sec:results}

\begin{figure}
    \centering
    \includegraphics[width=\linewidth]{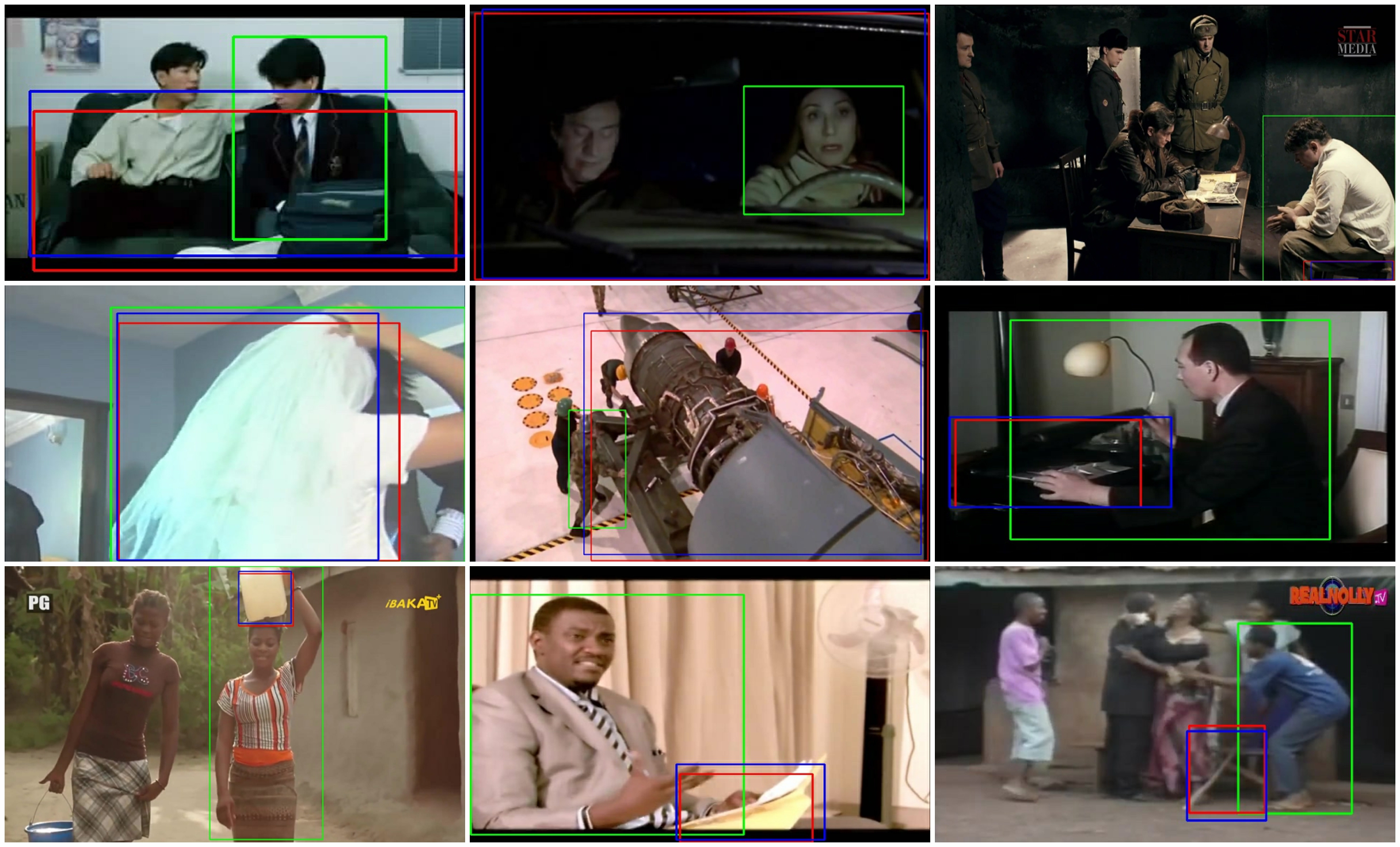}
    \caption{Some accurate predictions of CG-STVG. Green, blue, and red indicate the human box, GT object box, and predicted object box, respectively.}
    \label{fig:res-good}
\end{figure}

\begin{figure}
    \centering
    \includegraphics[width=\linewidth]{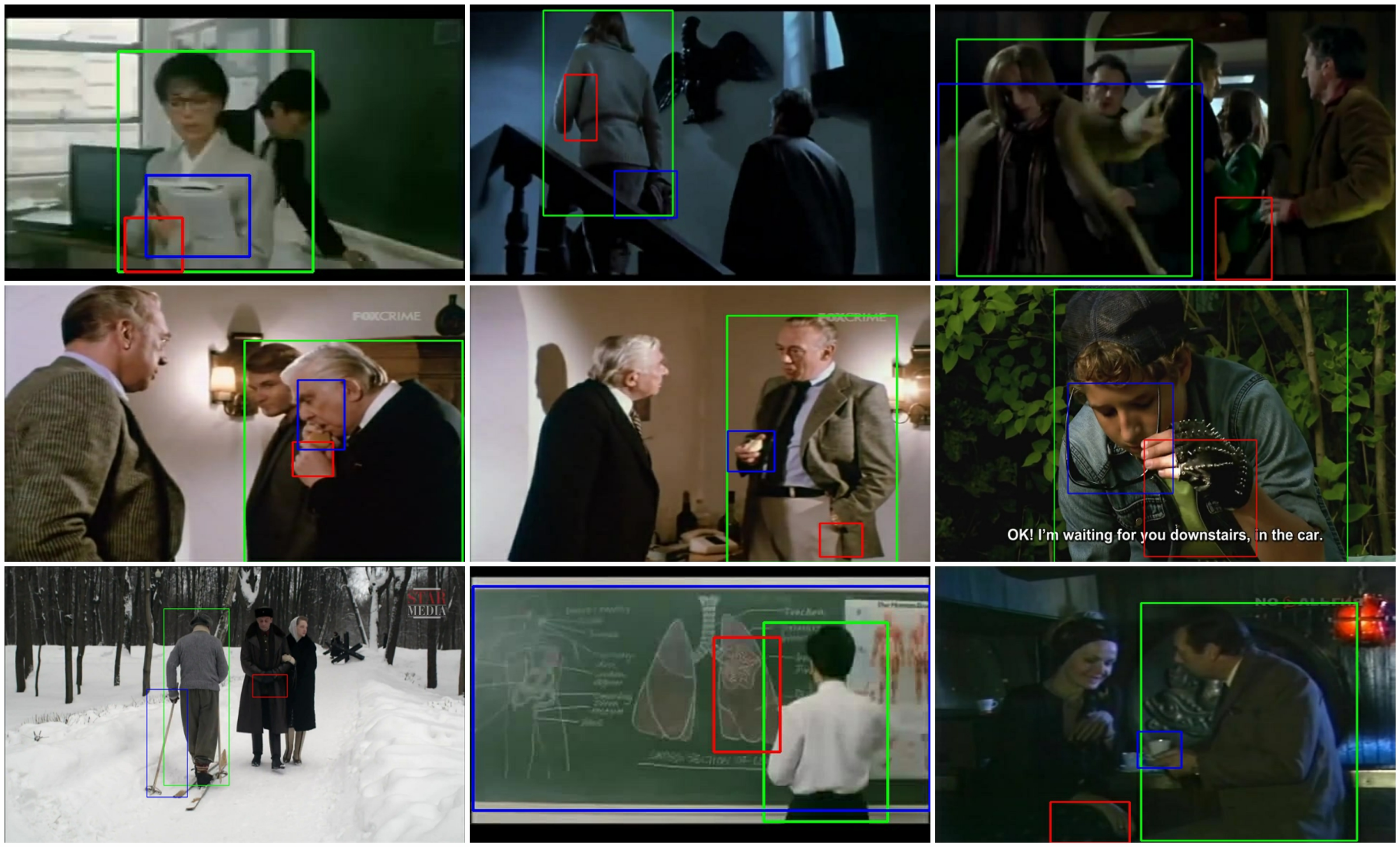}
    \caption{Some bad predictions of CG-STVG. Green, blue, and red indicate the human box, GT object box, and predicted object box, respectively.}
    \label{fig:res-bad}
\end{figure}

{\bf Baselines.} We show the state-of-the-art video-grounding baseline CG-STVG~\cite{cgstvg} results in the main paper. The video-grounding task is a more similar one to our GIO tasking setting, i.e., interacted object grounding, compared to Gaze's setting~\cite{gaze} and Grounding DINO's setting ~\cite{groundingdino} and others as listed in the Tab.~{2} in the main paper. Some CG-STVG visualization results are shown in Fig.~\ref{fig:res-good} and Fig.~\ref{fig:res-bad}.

{\bf More Ablations.} We have discussed different configurations of our 4D-QA. In the ablation section, we reported the 4D-QA w/o action with \textbf{19.04 mAP@0.5} and \textbf{27.30 mIoU$_w$}. By comparison, We conduct another experiment on Grounding DINO w/o action and get \textbf{13.25mAP@0.5} and \textbf{18.32 mIoU$_w$}, indicating that the interaction feature, or the semantic context, plays a crucial role in the grounding task.

\begin{figure}
    \centering
    \includegraphics[width=\linewidth]{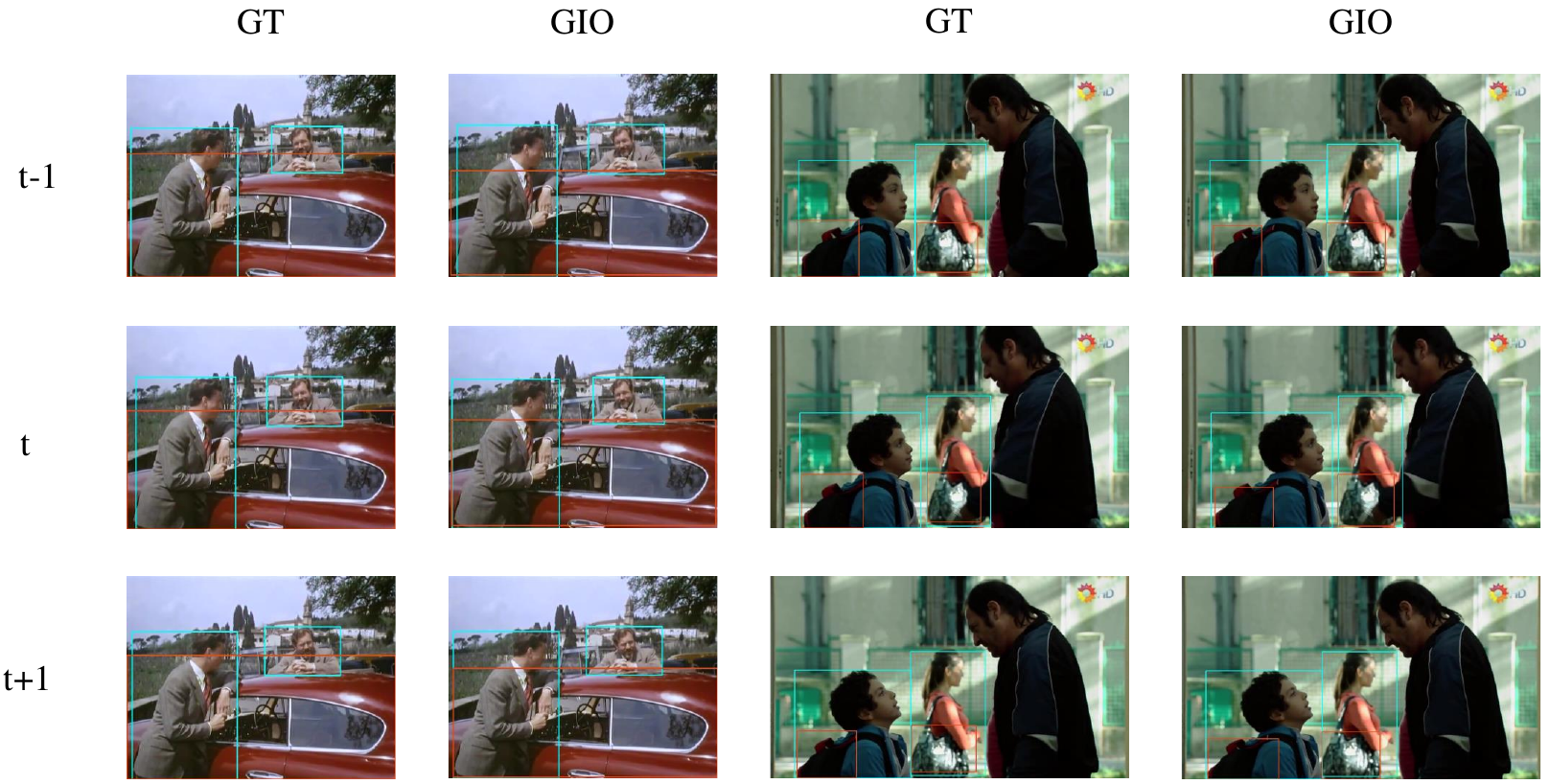}
    \caption{One frame with multiple HOIs results.}
    \label{fig:multi}
\end{figure}

{\bf More Comparisons in Single/Multiple HOI Scenarios.}
Our model performs well in both the single and multiple scenarios. For the 13k frames with multiple human-object pairs, 4D-QA provides 29.74 mIoU$_w$ and 23.52 mAP@0.5. For the 9k frames with one human-object pair, 4D-QA provides 29.59 mIoU$_w$ and 22.80 mAP@0.5. Some samples containing multiple pairs of HOI are shown in Fig.~\ref{fig:multi}.

\begin{figure*}[t]
    \centering
    \includegraphics[width=0.95\linewidth]{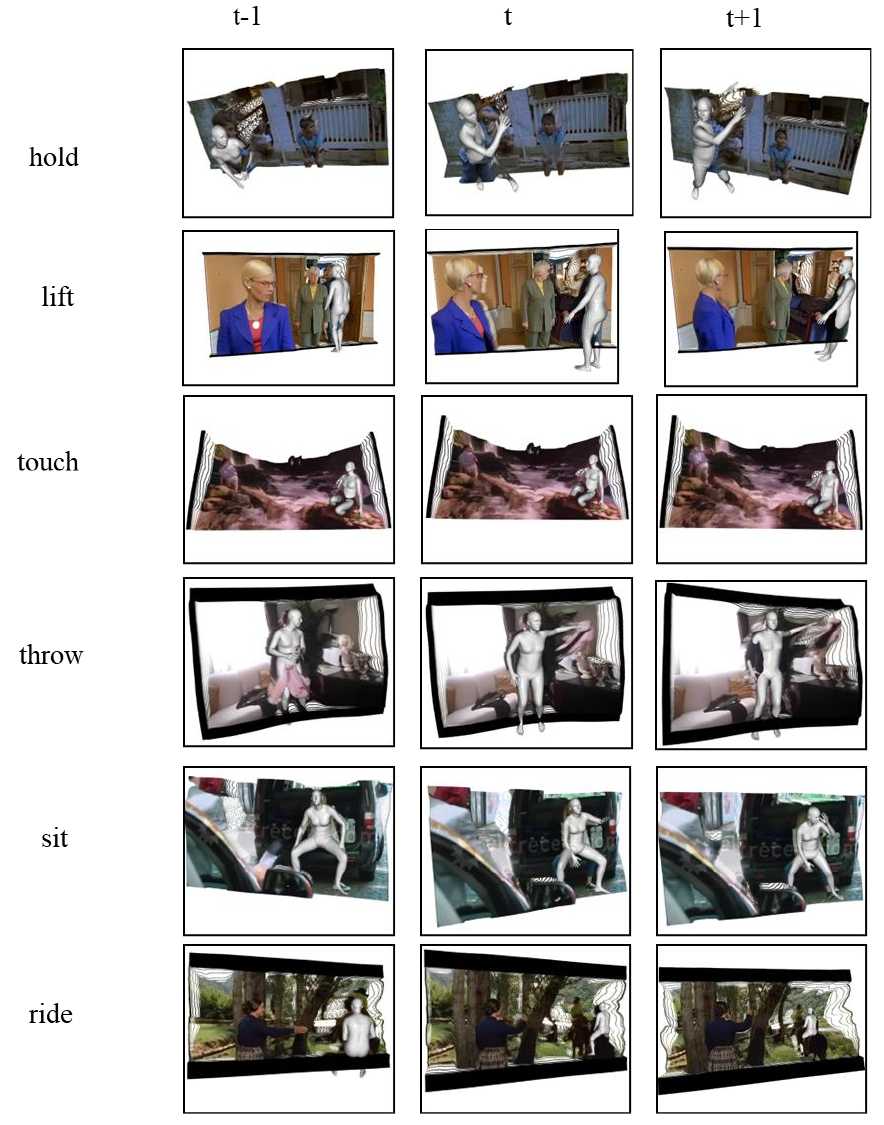}
    \caption{Visualization of 3D reconstructions from 4D-QA.}
    \vspace{-10px}
    \label{fig:3d}
\end{figure*}

\begin{figure*}[t]
    \centering
    \includegraphics[width=.99\linewidth]{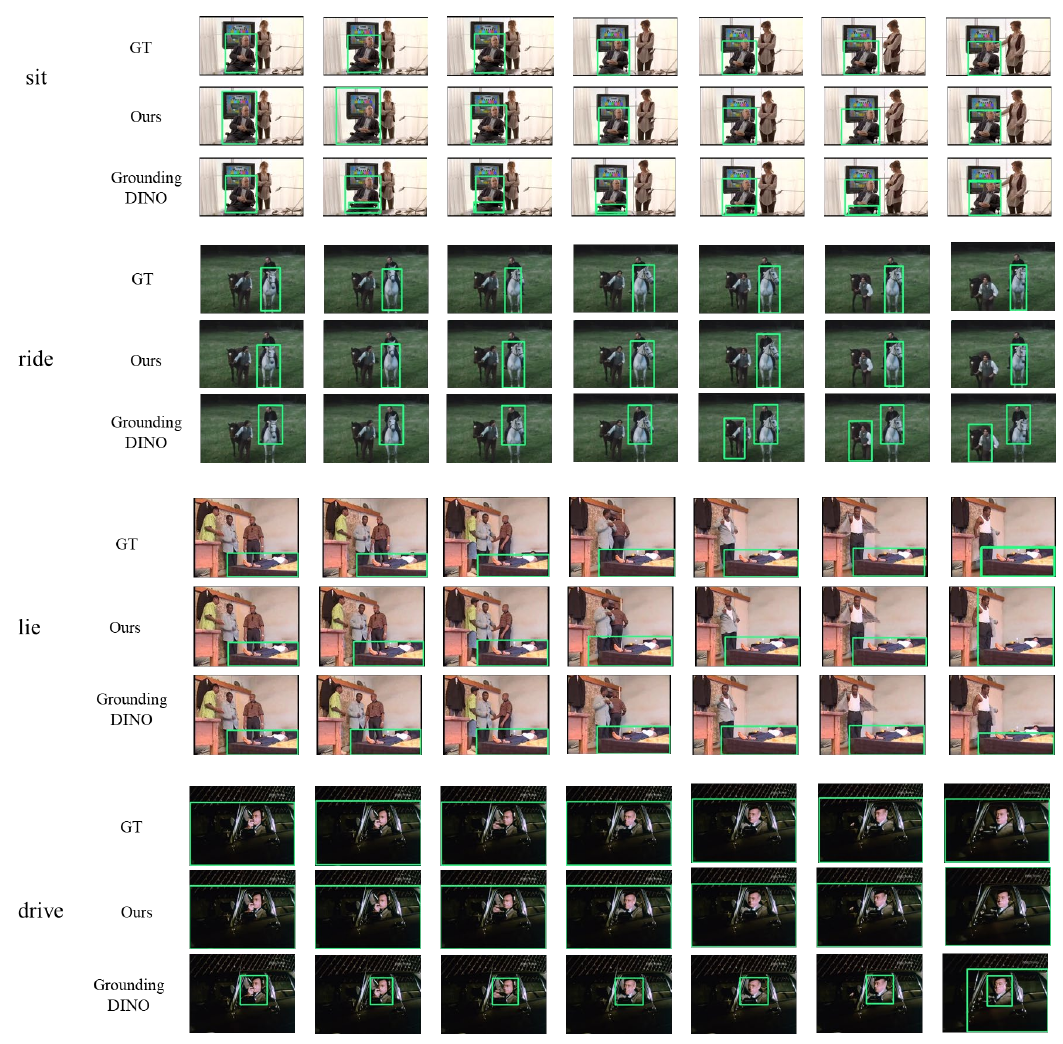}
    \caption{Visualization of grounding results.}
    \label{fig:grounding}
\end{figure*}

{\bf More Visualizations.} We also visualize some 3D reconstruction and grounding results of our method in Fig.~\ref{fig:3d} and Fig.~\ref{fig:grounding}. From the results, we can find that our method can recover the 4D scene changes and human-object interactions well, and 4D-QA also grounds different scales of objects robustly.
\section{Discussion}
\label{sec:discussion}
\noindent{\bf Limitations.} 
First, the current 4D-QA framework performs sub-optimally for small objects, which could be attributed to the granularity mismatch between object annotations and SAM-generated masks. We may try more advanced small object segmentation tools in future work.
Second, the 3D part of 4D-QA could be restricted by the sometimes over-flat depth estimation results adopted for scene reconstruction, due to the limitation of the depth estimators.

\noindent{\bf Future Explorations.} 
First, as revealed in Tab.~{5}, LLM-based methods unexpectedly provide relatively poor efficacy, demonstrating their potential weakness in interactiveness reasoning. Exploration of this would be a promising future work. 
Second, incorporating 3D features is critical for addressing occlusion and resolving spatial ambiguities in video analysis. These features have the potential to enhance a wide range of detection and perception models by serving as a "plug-in", thereby improving the comprehensiveness and accuracy of detection and analysis. Our findings demonstrate the feasibility of leveraging cross-modality features in detection tasks. Future research may focus on optimizing the extraction and utilization of 3D features to maximize their effectiveness.

\noindent{\bf Broader Impacts.}
A more advanced understanding of HOI could advance domestic health care, human-robot collaboration, etc. 
However, the potential abuse could result in an invasion of privacy.


\end{document}